\ifdefined\pdfminorversion
\fi
\documentclass{article} 
\usepackage[OT1]{fontenc} 
\usepackage{iclr_conference,times}
\usepackage{algorithm}
\usepackage{algorithmic}
\usepackage{wrapfig,needspace}
\usepackage{capt-of}
\usepackage{graphicx}
\usepackage{tikz}
\usepackage{booktabs}
\usepackage{multirow}
\usepackage{array,pifont}

\usepackage{amsmath,amsfonts,bm}

\def\1{\bm{1}}

\DeclareMathAlphabet{\mathsfit}{\encodingdefault}{\sfdefault}{m}{sl}
\SetMathAlphabet{\mathsfit}{bold}{\encodingdefault}{\sfdefault}{bx}{n}

\usepackage{hyperref}
\usepackage{url}
\usepackage{xcolor}
\usepackage{colortbl}
\definecolor{MGATeal}{HTML}{1F7A78}
\definecolor{MGAIndigo}{HTML}{4B2E83}
\definecolor{MGAOrange}{HTML}{D9660E}
\definecolor{MGAGreen}{HTML}{4C9A55}
\definecolor{MGAOrangeInk}{HTML}{A94F0B}
\definecolor{MGAAppendixInk}{HTML}{52627A}

\hypersetup{
    colorlinks=true,
    linkcolor=MGATeal,
    citecolor=MGAIndigo,
    urlcolor=MGATeal,
    filecolor=MGAIndigo,
    pdfborder={0 0 0}
}

\DeclareRobustCommand{\MGA}{\textcolor{MGATeal}{MGA}}
\DeclareRobustCommand{\MGAName}{\textcolor{MGATeal}{Manifold Generative Annealing}}
\DeclareRobustCommand{\MGAInitials}{\textcolor{MGATeal}{\textbf{M}anifold \textbf{G}enerative \textbf{A}nnealing}}

\DeclareRobustCommand{\visualref}[1]{%
    \begingroup\hypersetup{linkcolor=MGAOrangeInk}\ref{#1}\endgroup}
\DeclareRobustCommand{\appref}[1]{%
    \begingroup\hypersetup{linkcolor=MGAAppendixInk}\ref{#1}\endgroup}

\makeatletter
\let\mga@tagform@\tagform@
\renewcommand{\tagform@}[1]{%
    \begingroup\hypersetup{linkcolor=MGAOrangeInk}%
    \mga@tagform@{\textcolor{MGAOrangeInk}{#1}}\endgroup}
\renewcommand{\fnum@figure}{\textcolor{MGAOrangeInk}{\figurename~\thefigure}}
\renewcommand{\fnum@table}{\textcolor{MGAOrangeInk}{\tablename~\thetable}}
\renewcommand{\fnum@algorithm}{\textcolor{MGAOrangeInk}{\fname@algorithm~\thealgorithm}}
\makeatother

\title{Manifold Generative Annealing for Contact Constraint-aware Impedance Control with Reinforcement Learning Prior}

\author{Zhilin He \\
Carnegie Mellon University\\
\texttt{hectorh@andrew.cmu.edu}
\And
Guoquan Qi \\
Institute of Automation, \\
Chinese Academy of Sciences \\
\texttt{guoquan.qi@ia.ac.cn}
\And
Changliu Liu \\
Carnegie Mellon University \\
\texttt{cliu6@andrew.cmu.edu}
}

\iclrfinalcopy 
\begin{document}

\begingroup
\setlength{\tabcolsep}{2pt} 
\maketitle
\endgroup
\lhead{Preprint.} 

\begin{abstract}
Contact-rich robotic tasks require coordinated motion and impedance under uncertain interaction dynamics and physical constraints. While recent methods on learned policies provide useful interaction strategies, they may fail when contact conditions differ from those encountered during training. Model-based generative methods show promising receding-horizon trajectories optimization results, but often lack structured behavioral guidance and refinement that jointly accounts for constraint geometry and actual contact responses. To this end, we propose \textbf{\MGAName{} (\MGA{})}, a joint model-free and model-based generative framework for motion--impedance optimization. \MGA{} first organizes pretrained reinforcement learning policies into structured sequence proposals, retaining task-specific interaction strategies without committing to their execution. It then evaluates these proposals alongside exploratory samples through model-based rollouts for refinement or separate execution selection. Crucially, realization-aware manifold refinement aligns updates from the refinement bank with both task constraints and locally achievable physical responses, allowing learned experience to guide online adaptation under changing contact conditions. Experiments cover surface scanning, peg insertion, and humanoid pushing, with improved safety success rate of impedance control under nominal and sensing-shift conditions. Code: \href{https://github.com/hhhhzl/genedynamics}{\texttt{https://github.com/hhhhzl/genedynamics}}
\end{abstract}

\vspace{-0.2cm}
\section{Introduction}
\vspace{-0.2cm}
\label{sec:introduction}

Contact-rich tasks such as surface scanning, peg insertion, and whole-body pushing require robots to coordinate motion with compliant interaction while respecting physical constraints. Impedance learning and control provide an interface for regulating the relationship between motion and interaction forces~\citep{hogan1984impedance, buchli2011learning, martin2019variable}. Selecting these commands remains challenging due to multimodal interaction behaviors, uncertain physical responses, and safety-critical execution, requiring their joint optimization to sustain task progress under changing contact conditions~\citep{abu2020variable, tsuji2025survey, zhang2025safe}.

Safe Reinforcement Learning (RL) combines learned task policies with stiffness adaptation, recovery mechanisms, or action safeguards~\citep{zhao2022modelfree, zhang2024srl}, providing behavioral structure for contact-rich decisions. Beyond individual policy outputs, the same learned experience can define distributions over candidate behaviors. Model-free diffusion planners provide this generative extension by representing multimodal trajectory distributions, while constrained variants incorporate model-based projections and primal--dual sampling to enforce deployment-time constraints~\citep{romer2024diffusion,christopher2024constrained,zhang2025constrained,blanke2025sal}. These learned distributions can therefore supply task-informed candidates that reduce unguided search. However, they remain proposals rather than reliable plans because their relevance is inherited from the training distribution and may deteriorate as task or contact conditions change.

Model-Based Diffusion (MBD)~\citep{pan2024model} offers a way to revise sequences through their predicted consequences: it casts trajectory generation as optimization using objective-weighted dynamics rollouts without a learned denoising score. Constrained extensions~\citep{he2026, he2026motion} improve feasibility through Control Barrier Function (CBF)~\citep{ames2019control} or Convex Feasible Set (CFS)~\citep{liu2018convex} with adaptive scheduling. Building on this trajectory-optimization view, DIAL-MPC brings diffusion-style annealing to short-horizon model predictive control for online refinement~\citep{xue2025full, ye2025pegasusflow}. With a limited rollout budget, learned proposals can focus this search on useful contact strategies, and model-based evaluation can reassess them under current conditions. However, weighting promising candidates does not ensure that their combined sequence preserves nonlinear task relationships. Evaluation therefore leaves a second problem: how to correct the update it produces.

Constraint geometry provides local information for correcting such updates. ATACOM constructs safe actions in the tangent space of constraint manifolds~\citep{liu2022robot,liu2025safe}, while geometry-aware generative optimization uses active constraints to shape refinement and exploration~\citep{he20262go}. Geometric impedance control further exploits $\mathrm{SE}(3)$ structure in learned variable-impedance policies for contact-rich manipulation~\citep{seo2023contact}. Under compliant contact, jamming, or switching, however, displacement commands may produce deformation or force, and stiffness changes the local response. Refinement must therefore account for measured tracking and contact responses when defining the task residual. Shaping the update can reduce the normal response it introduces, but residuals already present or left after a finite step require a further correction. These two operations connect model-based reassessment to continued contact execution.

We propose \MGAInitials{} (\MGA{}), a short-horizon controller that organizes learned behavioral structure, model-based reassessment, and feedback-conditioned geometric correction in the same motion--impedance sequence space. Rollout weighting determines which candidates inform the update. Geometry shaping adjusts its direction. Local residual correction then revises the updated sequence. The proposal and evaluation stages supply task-informed candidates for refinement or separate execution selection. Our \textcolor{MGAGreen}{\textbf{C}}ontributions are three-fold:
\vspace{-0.2cm}
\begin{itemize}
\item \textcolor{MGAGreen}{\textbf{C1}:} We organize model-free learned action policies into structured short-horizon motion--impedance proposals, supplying task-specific interaction behaviors alongside Gaussian exploration in a common sequence space.
\item \textcolor{MGAGreen}{\textbf{C2:}} We integrate these candidates with model-based generative annealing (online optimization) through common closed-loop evaluation, reassessing their relevance under current task and contact conditions through bank weighting or separate expert selection.
\item \textcolor{MGAGreen}{\textbf{C3:}} We develop realization-aware manifold refinement that adjusts update directions and corrects remaining task residuals. Contact feedback updates the task references, while response reliability controls the refinement strength. We analyze how the normalized core suppresses encoded normal responses and reduces local residuals.
\end{itemize}

\vspace{-0.2cm}
\section{Preliminaries}
\vspace{-0.2cm}
In this section, we cover the relevant background. Related Works detailed in Appendix~\appref{app:related_works}. We use a subscript $k$ for generative steps, a superscript $t$ for physical time, and a subscript $m$ for samples (e.g., $u_{k,m}^{t}$ denotes sample $m$'s control input at physical time $t$ and generative step $k$).

\label{sec:prelim}

\vspace{-0.2cm}
\subsection{Constrained Lower-Control Sequence Optimization}
\vspace{-0.2cm}
\label{subsec:prelim_lower_control}
\label{subsec:prelim_position_stiffness}
\label{subsec:prelim_constraint_manifold}

We consider a robot with state $s^t\in\mathcal S$ and generalized
lower-control input $u^t\in\mathcal U$ over a finite horizon $T$. Rather than sampling an independent
state trajectory, we optimize the finite lower-control sequence
\begin{math}
U=(u^0,\ldots,u^{T-1})\in\mathcal U^T.
\end{math}
At each step, a task-owned lower-level controller $\pi$ maps the decision to the
physical robot action. Together with the dynamics model $f$, this controller
induces a closed-loop rollout. The resulting constrained lower-control planning
problem is
{%
\setlength{\abovedisplayskip}{2.5pt}
\setlength{\belowdisplayskip}{2.5pt}
\setlength{\abovedisplayshortskip}{1.5pt}
\setlength{\belowdisplayshortskip}{1.5pt}
\begin{equation}
\begin{aligned}
& U^\star  \in
\arg\min_{U}
\quad J(U;s^0,c)
:=
\ell_T(s^T;c)
+
\sum_{t=0}^{T-1}\ell(s^t,u^t;c)
\\
\text{\rm s.t.}\quad
&
a^t=\pi(s^t,u^t;c), \quad
s^{t+1} = f(s^t,a^t;c), \quad
h(s^t,u^t;c)=0, \quad
g(s^t,u^t;c)\leq 0.
\end{aligned}
\label{eq:lower_control_ocp}
\end{equation}
}%
Context $c$ contains goals, geometry, contact mode, object or terrain
state, and model parameters. The interface covers torque control with
$\pi$ as the identity, impedance control, and whole-body control.

For contact-rich tasks, we instantiate Eq.~\eqref{eq:lower_control_ocp} with the
position--stiffness primitive
\begin{math}
u^t=(r^t,S^t,\nu^t),
\label{eq:position_stiffness_u}
\end{math}
where $r^t$ is a task, contact, or whole-body reference and $\nu^t$ contains
feedforward quantities such as desired force, velocity, or footsteps~\citep{hogan1984impedance, buchli2011learning, martin2019variable, zhang2024srl}.
Here $S^t=(S^t)^\top\in\mathbb R^{d\times d}$ is a log-stiffness decision
matrix, whereas $s^t$ denotes the robot state. The matrix exponential gives
the physical stiffness
\begin{math}
K^t=\exp(S^t)\in\mathbb{S}_{++}^{d}.
\label{eq:stiffness_log_param}
\end{math}
Thus positive-definite stiffness is feasible by construction
\citep{arsigny2007geometric}, while $h$ and $g$ impose the remaining constraints.
Equalities $h(\cdot)$ encode reliable contact or kinematic consistency. Inequalities
$g(\cdot)$ encode limits such as friction, force, collision, joint range, balance, or
tipping margins. To express these horizon constraints as a common zero-level
set, we define the residual map
{%
\setlength{\abovedisplayskip}{2.5pt}
\setlength{\belowdisplayskip}{2.5pt}
\setlength{\abovedisplayshortskip}{1.5pt}
\setlength{\belowdisplayshortskip}{1.5pt}
\begin{equation}
C(U,\mu;s^0,c)
=\left[
\begin{array}{c}
h(s^t,u^t;c)\\
g(s^t,u^t;c)+\tfrac12\mu^t\odot\mu^t
\end{array}
\right]_{t=0}^{T-1}.
\label{eq:constraint_vector}
\end{equation}
}%
Thus, $C(U,\mu;s^0,c)=0$ represents $h=0$ and $g\leq0$ through compatible
real variables $\mu$, whose squared entries supply nonnegative slack~\citep{he2026}.
This defines the hard feasible set in augmented coordinates~\citep{abudakka2020geometry, liu2022robot}
\begin{equation}
\mathcal M(s^0,c)
=\left\{(U,\mu)\ \middle|\;
C(U,\mu;s^0,c)=0\right\}.
\label{eq:control_manifold}
\end{equation}
Let $\mathcal I_{\rm mode}$ index discrete contact choices, such as active
object faces or stance feet. Each $b$ defines a local branch
$\mathcal M_b$, giving the union
\begin{math}
\mathcal M(s^0,c)=\bigcup_{b\in\mathcal I_{\rm mode}}\mathcal M_b(s^0,c).
\label{eq:hybrid_union_manifold}
\end{math}
On a fixed regular branch, let $a$ denote the set of active inequality row indices, where
$g_i(s^t,u^t;c)=0$. With $a$ fixed, $C_a(U)$ collects all equality rows and
these active rows after eliminating their slack. Let
$J_a(U)=\partial C_a/\partial U$ be its Jacobian through the rollout, with
fixed $(s^0,c)$ suppressed. At a feasible point,
$J_a\delta U=0$ defines this branch's tangent space. For a
positive-definite sequence metric $G(U)$ with
$\lVert\delta U\rVert_G^2=\delta U^\top G\delta U$, the $G$-orthogonal
tangent projector $P_a^G$ is~\citep{dietrich2015overview}
{%
\setlength{\abovedisplayskip}{2.5pt}
\setlength{\belowdisplayskip}{2.5pt}
\setlength{\abovedisplayshortskip}{1.5pt}
\setlength{\belowdisplayshortskip}{1.5pt}
\begin{equation}
P_a^{G}(U)
=I-G^{-1}J_a^\top
\left(J_aG^{-1}J_a^\top\right)^\dagger J_a.
\label{eq:metric_tangent_projection}
\end{equation}
}%
where $\dagger$ denotes the Moore--Penrose pseudoinverse~\citep{penrose1955generalized}. This local map provides
a template for realization-aware refinement, without guaranteeing finite-step feasibility.
To combine learned proposals with model-based evaluation, we next express
sequence selection as probabilistic inference, assigning greater weight to
low-cost, feasible candidates.

\subsection{Joint Model-free Model-Based Generative Inference}
\label{subsec:prelim_joint_inference}

The constrained cost in Eq.~\eqref{eq:lower_control_ocp} defines the formal
model-based compatibility factor
\begin{equation}
\mathcal L(U;s^0,c)
=
\exp[-J(U;s^0,c)/\beta]\,
\mathbf 1\!\left\{\exists\mu:\ C(U,\mu;s^0,c)=0\right\},
\label{eq:model_based_likelihood}
\end{equation}
where $\beta>0$ is a temperature and the indicator enforces the existence of
compatible slack variables in Eq.~\eqref{eq:control_manifold}.  Thus,
lower-cost feasible sequences receive higher model-based compatibility.

A model-free component provides a complementary trajectory distribution
$Q_\psi(\cdot\mid s^0,c)$.  A common instantiation is a state-conditioned RL
policy trained with
{%
\setlength{\abovedisplayskip}{2.5pt}
\setlength{\belowdisplayskip}{2.5pt}
\setlength{\abovedisplayshortskip}{1.5pt}
\setlength{\belowdisplayshortskip}{1.5pt}
\begin{equation}
\psi^\star\in\arg\max_\psi\quad
\mathbb E_{\pi_\psi}\!\left[
\sum_{t=0}^{\infty}\gamma^t\mathcal R(s^t,u^t;c)
\right],
\label{eq:rl_objective}
\end{equation}
}%
where $\gamma\in[0,1)$ and $\mathcal R$ is the policy-training reward.
Given $s^0$, the policy induces a sequence distribution through the
closed-loop rollout in Eq.~\eqref{eq:lower_control_ocp}:
\begin{equation}
u^t\sim\pi_\psi(\cdot\mid o^t,c), \quad
U=(u^0,\ldots,u^{T-1})\sim Q_\psi(\cdot\mid s^0,c),
\label{eq:closed_loop_rollout}
\end{equation}
where $o^t$ is the policy observation associated with $s^t$.  More generally,
$Q_\psi$ can be induced by any learned policy or trajectory model, including
an implicit or deterministic generator.  Its training objective need not
coincide with the test-time planning objective: the model-free component
supplies behavioral structure, whereas the model-based factor supplies
test-time evaluation.

Joint model-free model-based inference reweights this learned trajectory
distribution using model-based evidence~\citep{jung2025joint}.  Directly using
$\mathcal L$ gives a zero normalizer when the equality constraints define a
lower-dimensional feasible set and $Q_\psi$ has an ambient density.  To retain
nonzero weights for mildly infeasible candidates, we use the softened cost
\begin{math}
\widetilde J(U;s^0,c)
=J(U;s^0,c)+\mathcal V(U;s^0,c),
\end{math}
where the scalar penalty $\mathcal V\geq0$ measures task-selected constraint violations along
the rollout.  This penalty evaluates candidates
without enforcing hard feasibility or replacing geometric refinement.
Following Boltzmann distribution~\citep{watson2020stochastic, watson2023inferring}, the resulting joint-inference target is
{%
\setlength{\abovedisplayskip}{2.5pt}
\setlength{\belowdisplayskip}{2.5pt}
\setlength{\abovedisplayshortskip}{1.5pt}
\setlength{\belowdisplayshortskip}{1.5pt}
\begin{equation}
P_\beta(dU\mid s^0,c)
\propto
\exp[-\widetilde J(U;s^0,c)/\beta]\,
Q_\psi(dU\mid s^0,c),
\label{eq:soft_target_distribution}
\end{equation}
}%
assuming a finite, nonzero normalizer.  This is a prior-weighted surrogate for
Eq.~\eqref{eq:lower_control_ocp}.  For samples from $Q_\psi$, weighting requires
only rollout costs, not a prior density or score.  Reweighting cannot expand the support of
$Q_\psi$, leaving a deterministic horizon unchanged. \MGA{} therefore
complements learned proposals with local Gaussian exploration.  We next
introduce a generative sampling view that separates clean-sequence estimation
from the reverse transport that updates the sequence.

\subsection{Generative Models for Trajectory Sampling}
\label{subsec:prelim_diffusion_control}

Take $P_\beta$ in Eq.~\eqref{eq:soft_target_distribution} as the clean sequence
measure. Generative sampling refines a noisy sequence $U_K$ toward a clean sequence
$U_0$.  In normalized control coordinates, we define a Gaussian probability path by
\begin{math}
U_k=\sqrt{\bar\alpha_k}U_0+\sqrt{1-\bar\alpha_k}\,\epsilon,
\end{math}
where $\epsilon\sim\mathcal N(0,I)$,
$\bar\alpha_k=\prod_{j=1}^{k}\alpha_j$ with $\alpha_j\in(0,1)$,
$\bar\alpha_0=1$, and $\bar\alpha_K\approx0$. For $k\geq1$, the forward kernel $q_k(U_k\mid U_0)=\mathcal N(U_k;\sqrt{\bar\alpha_k}U_0,
(1-\bar\alpha_k)I)$ is the conditional density of the noisy sequence given
the clean one. The noisy marginal is $p_k(U_k)
=\int q_k(U_k\mid U_0)P_\beta(dU_0)$. The clean conditional law is
{%
\setlength{\abovedisplayskip}{2.5pt}
\setlength{\belowdisplayskip}{2.5pt}
\setlength{\abovedisplayshortskip}{1.5pt}
\setlength{\belowdisplayshortskip}{1.5pt}
\begin{equation}
\begin{aligned}
P_{0\mid k}(dU_0\mid U_k)
\propto q_k(U_k\mid U_0)
\exp[-\widetilde J(U_0)/\beta]\,Q_\psi(dU_0).
\end{aligned}
\label{eq:joint_clean_conditional}
\end{equation}
}%
Let $\bar U_k:=\mathbb E_{P_{0\mid k}}[U_0\mid U_k]$ be the conditional clean
mean. The Gaussian Tweedie identity~\citep{efron2011tweedie} relates it to the noisy score through
{%
\setlength{\abovedisplayskip}{2.5pt}
\setlength{\belowdisplayskip}{2.5pt}
\setlength{\abovedisplayshortskip}{1.5pt}
\setlength{\belowdisplayshortskip}{1.5pt}
\begin{equation}
\mathcal S_k(U_k;\widehat U_0)
=-\frac{U_k}{1-\bar\alpha_k}
+\frac{\sqrt{\bar\alpha_k}}{1-\bar\alpha_k}\widehat U_0,
\label{eq:mb_score_U}
\end{equation}
}%
where $\widehat U_0$ is a generic estimate of the clean sequence $U_0$. The exact score satisfies
$\nabla_{U_k}\log p_k(U_k)=\mathcal S_k(U_k;\bar U_k)$.
A learned denoiser can approximate $\bar U_k$, while model-based methods use
rollout evaluations to construct a clean estimate.

For the Gaussian-only diffusion special case, a flat reference replaces $Q_\psi$.
Normalizing $q_k(U_k\mid U_0)$ over $U_0$ gives the proposal used in
model-based sampling~\citep{pan2024model,xue2025full}:
{%
\setlength{\abovedisplayskip}{2.5pt}
\setlength{\belowdisplayskip}{2.5pt}
\setlength{\abovedisplayshortskip}{1.5pt}
\setlength{\belowdisplayshortskip}{1.5pt}
\begin{equation}
\begin{aligned}
\widetilde U_{k,m}\sim\Omega_k(U_k)
:=\mathcal N\!\left(
\frac{U_k}{\sqrt{\bar\alpha_k}},
\left(\frac{1}{\bar\alpha_k}-1\right)I
\right),\quad
\widehat{\bar U}_k
=\frac{\sum_{m=1}^{M}w_{k,m}\widetilde U_{k,m}}
{\sum_{m=1}^{M}w_{k,m}}.
\end{aligned}
\label{eq:clean_proposal_U}
\end{equation}
}%
where $w_{k,m}=e^{-\widetilde J(\widetilde U_{k,m})/\beta}$. Each candidate is evaluated through Eq.~\eqref{eq:lower_control_ocp}.
These cost-only weights rely on the flat reference. Arbitrary learned
proposals do not yield the same cancellation.

\vspace{-0.1cm}
The clean estimate $\widehat{\bar U}_k$, or its corresponding score, can drive
different reverse transports, including stochastic diffusion, DDIM, and
probability-flow updates~\citep{song2020denoising,song2021score}. We write these as
{%
\setlength{\abovedisplayskip}{2.5pt}
\setlength{\belowdisplayskip}{2.5pt}
\setlength{\abovedisplayshortskip}{1.5pt}
\setlength{\belowdisplayshortskip}{1.5pt}
\begin{equation}
U_{k-1}=\mathcal T_k(U_k,\widehat{\bar U}_k,z_k),\quad
z_k\sim\mathcal N(0,I),
\label{eq:generative_transport}
\end{equation}
}%
where $\mathcal T_k$ specifies the transport (DDPM, DDIM, Flow-like), and deterministic updates omit $z_k$.
The question now becomes how to construct structured proposals and refine their clean moment
using realization geometry before applying the chosen transport.

\vspace{-0.2cm}
\section{Methodology}
\label{sec:method}
\vspace{-0.2cm}

\MGA{} couples three steps in motion--impedance sequence space.
(\textcolor{MGAGreen}{\textbf{1}}) Learned policies and Gaussian exploration supply candidate
behaviors. (\textcolor{MGAGreen}{\textbf{2}}) Common closed-loop rollouts reassess candidates
for refinement or separate execution selection. (\textcolor{MGAGreen}{\textbf{3}}) Geometric refinement coordinates the weighted bank update
with task residuals under current feedback. Directional shaping reduces the
update's encoded normal response. Local residual correction (LRC) then
corrects residuals remaining afterward.

\begin{figure}[!t]
    \centering
    \includegraphics[width=\linewidth,trim=80bp 25bp 0bp 0bp,clip]{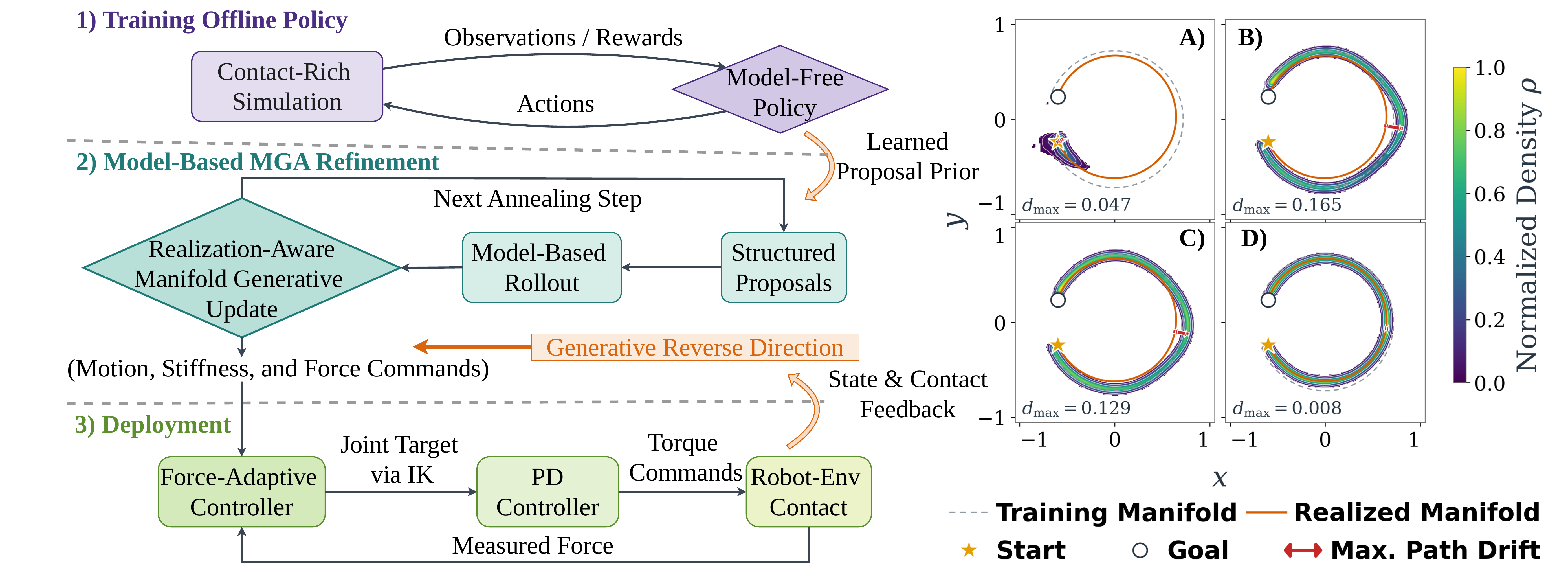}
    \caption{Left: Overview of \MGA{}: (1) offline policy training, (2) model-based refinement,
    and (3) force-adaptive deployment with feedback.
    Right: an open-circle start-to-goal toy: (A) Gaussian-initialized refinement,
    (B) RL prior, (C) \MGA{} w/o LRC, and (D) \MGA{}.
    The prior supplies path structure. LRC reduces maximum median-path drift:
    $0.165$ (B), $0.129$ (C), $0.008$ (D).
    Arrows mark full-path maxima of distance to the realized manifold. C/D differ only in LRC.}
    \label{fig:mga_overview}
    \vspace{-0.3cm}
\end{figure}

\vspace{-0.2cm}
\subsection{RL Prior-guided Structural Proposals}
\label{subsec:method_rl_prior}
\vspace{-0.2cm}

The policy supplies short-horizon interaction behaviors.
Gaussian proposals explore local alternatives. At reverse step $k$, they
are centered on the clean sequence
$\check U_k:=U_k/\sqrt{\bar\alpha_k}$ under the Gaussian path in
Section~\ref{subsec:prelim_diffusion_control}.
Policy rollouts follow $Q_\psi$. In solver-node coordinates $U$, rollouts and
residuals include fixed interpolation (Appendix~\appref{app:proposal_implementation}).
The refinement bank is
\begin{equation}
\mathcal Q_k
=
\{\widetilde U_{k,m}^{\rm g}\}_{m=1}^{M_{\rm g}}
\uplus\{\widetilde U_{k,m}^{\psi}\}_{m=1}^{M_\psi},
\quad M_{\rm g}+M_\psi=M,\quad M_{\rm g}>0.
\label{eq:fixed_budget_proposal_mixture}
\end{equation}
Here $M$ counts refinement candidates. Replacement exchanges Gaussian
for learned slots at fixed $M$. Additive sets $M_\psi=0$ and compares
separate policy horizons with the refined plan at execution selection.
Learned horizons are drawn once per solve. Replacement slots are reused
across reverse steps, while Gaussian slots are resampled.
The anchor set $\mathcal A_k^{\rm anc}$
contains $\check U_k$ and, when used, a deterministic policy rollout.
Appending them gives $N_k=M+|\mathcal A_k^{\rm anc}|$ slots $\widetilde U_{k,m}$.

\vspace{-0.4cm}
\paragraph{Model-based Weighting.}
The candidate sources suggest behaviors, but their usefulness depends on
current contact conditions. We evaluate bank candidates through
the same closed-loop model in Eq.~\eqref{eq:lower_control_ocp}. The softened
objective $\widetilde J$ weights them into a clean sequence estimate:
{%
\setlength{\abovedisplayskip}{2.5pt}
\setlength{\belowdisplayskip}{2.5pt}
\setlength{\abovedisplayshortskip}{1.5pt}
\setlength{\belowdisplayshortskip}{1.5pt}
\begin{equation}
\bar w_{k,m}
=\operatorname{softmax}_m\!\left(
-\frac{\widetilde J(\widetilde U_{k,m})}{\beta_k}
\right), \quad
\widehat{\bar U}_k
=\sum_{m=1}^{N_k}\bar w_{k,m}\widetilde U_{k,m}.
\label{eq:method_clean_mean}
\end{equation}
}%
Here $\beta_k>0$ is the cost-scaled temperature. Rollout costs determine each
candidate's contribution without requiring a learned density or score.
This finite-bank weighting follows the principle of
Eq.~\eqref{eq:soft_target_distribution}, without sampling exactly from
$P_\beta$. The proposed update $d_k:=\widehat{\bar U}_k-\check U_k$
moves the current clean-coordinate sequence toward the weighted estimate.
Even low-cost candidates can yield a weighted combination that disrupts
nonlinear task relationships. We therefore shape this update's direction
and correct the resulting sequence's task residual.

\subsection{Realization-Aware Manifold Refinement}
\label{subsec:method_realization_geometry}
\label{subsec:method_metric}
\label{subsec:method_reverse_update}

The residual $C^{\rm real}(U)$ measures a sequence's deviation from task
references. Where enabled, tracking and contact feedback adjust these
references using a local response estimate. This correction
$\Delta^{\rm real}$ remains fixed during the solve. The local zero-level set is
\begin{math}
\mathcal M^{\rm real}:=\{U:C^{\rm real}(U)=0\}.
\end{math}
Setting $\Delta^{\rm real}=0$ recovers the nominal task references.
This residual encodes selected task relationships, distinct from the full
hard-feasibility residual in Eq.~\eqref{eq:constraint_vector}.
Task-specific definitions and feedback corrections are given in
Appendix~\appref{app:realization_maps}.

We first shape the proposed update $d_k$. Let
$\mathsf J(U):=\partial C^{\rm real}(U)/\partial U$ be the residual Jacobian.
We construct $A_k$ from node blocks of the residual-energy gradient
(Appendix~\appref{app:geometry_implementation}), so $A_kd_k$ measures the
update along encoded normal directions. For a metric $G_k\succ0$ and
regularization $\lambda_k>0$, the filter below gives the shaped update
$\mathsf M_kd_k$, the updated clean estimate, and its reverse field:
\begin{equation}
\begin{aligned}
\mathsf M_k
&:=I-G_k^{-1}A_k^{\top}
\left(\lambda_k I+A_kG_k^{-1}A_k^{\top}\right)^{-1}A_k,\\
\bar U_{\mathcal M,k}&:=\check U_k+\mathsf M_kd_k,
\quad \widehat S_{\mathcal M,k}:=\mathcal S_k(U_k;\bar U_{\mathcal M,k}).
\end{aligned}
\label{eq:geometry_shaped_score}
\end{equation}
The filter preserves directions $z$ with $A_kz=0$ and attenuates
the encoded response $A_kd_k$. Smaller $\lambda_k$ suppresses the encoded
responses more strongly, while larger values retain more of $d_k$.
At a regular point of $\mathcal M^{\rm real}$, choosing
$A_k=\mathsf J(\check U_k)$ and taking $\lambda_k\downarrow0$ recovers
the metric tangent map in Eq.~\eqref{eq:metric_tangent_projection}.
We use $G_k=I$ in normalized motion--impedance coordinates.
Log-stiffness coordinates preserve positive definiteness, and rollout
evaluation accounts for motion--impedance coupling. Task-dependent gain
sets the refinement strength. Where enabled, response reliability also
modulates the change to the rollout-weighted update
(Appendix~\appref{app:geometry_implementation}).

We next carry the shaped estimate to reverse step $k-1$ using the
MBD-style transport in Eq.~\eqref{eq:generative_transport}:
{%
\setlength{\abovedisplayskip}{2.5pt}
\setlength{\belowdisplayskip}{2.5pt}
\setlength{\abovedisplayshortskip}{1.5pt}
\setlength{\belowdisplayshortskip}{1.5pt}
\begin{equation}
\begin{aligned}
U_{k-1}^{\rm raw}
=\frac{1}{\sqrt{\alpha_k}}
\left(U_k+(1-\bar\alpha_k)\widehat S_{\mathcal M,k}\right), \quad
\frac{U_{k-1}^{\rm raw}}{\sqrt{\bar\alpha_{k-1}}}
=\bar U_{\mathcal M,k}.
\end{aligned}
\label{eq:ddpm_control_update}
\end{equation}
}%
The identity on the right follows from
$\bar\alpha_k=\alpha_k\bar\alpha_{k-1}$: transport changes the reverse-step
coordinates while preserving the shaped clean estimate.

\vspace{-0.3cm}
\paragraph{Local Residual Correction (LRC).}
Shaping controls the update direction, but existing residuals and finite-step
errors can remain. LRC therefore corrects the transported clean sequence
$V_k:=U_{k-1}^{\rm raw}/\sqrt{\bar\alpha_{k-1}}=\bar U_{\mathcal M,k}$
using a damped Jacobian step:
{%
\setlength{\abovedisplayskip}{2.5pt}
\setlength{\belowdisplayskip}{2.5pt}
\setlength{\abovedisplayshortskip}{1.5pt}
\setlength{\belowdisplayshortskip}{1.5pt}
\begin{equation}
V_k^+=V_k-\eta\mathsf J(V_k)^\top
\left[\mathsf J(V_k)\mathsf J(V_k)^\top+\varepsilon I\right]^{-1}C^{\rm real}(V_k),
\quad U_{k-1}=\sqrt{\bar\alpha_{k-1}}V_k^+,
\label{eq:local_residual_correction}
\end{equation}
}%
Here $\eta>0$ sets the correction gain and $\varepsilon>0$ regularizes
the linear solve. The Jacobian-based term maps the remaining residual
$C^{\rm real}(V_k)$ into a sequence correction. LRC evaluates the full
Jacobian at $V_k$, whereas shaping uses $A_k$ at $\check U_k$.
This local correction is neither an exact projection nor a hard-feasibility
guarantee. Implementation details appear in
Appendix~\appref{app:proposal_implementation}.

\vspace{-0.2cm}
\subsection{Receding-Horizon Execution}
\label{subsec:method_summary}
\vspace{-0.2cm}

\begin{wrapfigure}{r}{0.61\linewidth}
\vspace{-\intextsep}
\input{tex/algorithm}
\end{wrapfigure}
\noindent
Algorithm~\visualref{alg:mga} summarizes one replanning cycle.
\MGA{} initializes from the shifted plan, optionally mixed with a deterministic
policy rollout, and fixes $\Delta^{\rm real}$.
Each reverse step evaluates the refinement bank. The resulting $d_k$
undergoes directional shaping, transport, and LRC. The algorithm presents
the normalized geometric core. Gain, reliability modulation, and
localization are specified in
Appendix~\appref{app:proposal_implementation}. After refinement, execution
acceptance, when enabled, re-evaluates the candidate against the task-defined
incumbent. Additive experts then compete through separate evaluation and
acceptance. The lower-level controller executes the first primitive,
and the remaining horizon is shifted for the next warm start.
\par
\WFclear

\vspace{-0.2cm}
\section{Theoretical Analysis}
\label{sec:theory}
\vspace{-0.2cm}
We summarize the main theoretical properties of \MGA{} and defer detailed
proofs to Appendix~\appref{app:proofs}. The analysis isolates the effects of
model-based reweighting, geometric refinement, and local residual correction.
We condition on the evaluated batch at reverse step $k$, with finite costs and
$\beta_k>0$, and fix $A_k$, $G_k\succ0$, and $\lambda_k>0$.
The results concern the normalized update before task-dependent modulation.
We use $\lVert x\rVert_{G_k}^2=x^\top G_kx$.
\vspace{-0.3cm}
\paragraph{Lemma 1 (Finite-bank Model-based Reweighting).}
Let $j_m:=\widetilde J(\widetilde U_{k,m})$ and let $\Delta_{N_k}$ be the
probability simplex over the $N_k$ evaluated candidates. The weights in
Eq.~\eqref{eq:method_clean_mean} uniquely solve
{%
\setlength{\abovedisplayskip}{2.5pt}
\setlength{\belowdisplayskip}{2.5pt}
\setlength{\abovedisplayshortskip}{1.5pt}
\setlength{\belowdisplayshortskip}{1.5pt}
\begin{equation}
\bar w_k=\arg\min_{w\in\Delta_{N_k}}
\left\{\sum_{m=1}^{N_k}w_mj_m
+\beta_kD_{\mathrm{KL}}\!\left(w\middle\|\frac{\mathbf 1}{N_k}\right)\right\}.
\label{eq:finite_bank_reweighting}
\end{equation}
}%
Moreover, $\sum_m\bar w_{k,m}j_m\leq\min_m j_m+\beta_k\log N_k$.
Thus model-based weighting concentrates the evaluated distribution toward
lower-cost candidates, regardless of their proposal source. This statement
concerns weighted candidate costs, not the cost or feasibility of their
ambient mean.

\begin{figure}[!t]
  \centering
    \caption{The environments are (a) Surface Scanning over rigid (purple) and soft (green) surfaces with varied geometries and hybrid hard--soft patterns, (b) Peg Insertion into a fixed socket under nominal conditions (ID), perturbed socket poses (Pose-OOD), and biased force/torque sensing (Sensing-OOD), and (c) Humanoid Push for constant $15$-N force regulation, fixed-stance pushing to the green line, and box unjamming with yaw correction.}
    \label{fig:environments}
    \begin{minipage}[t]{0.32\linewidth}
        \centering
        \includegraphics[width=\linewidth]{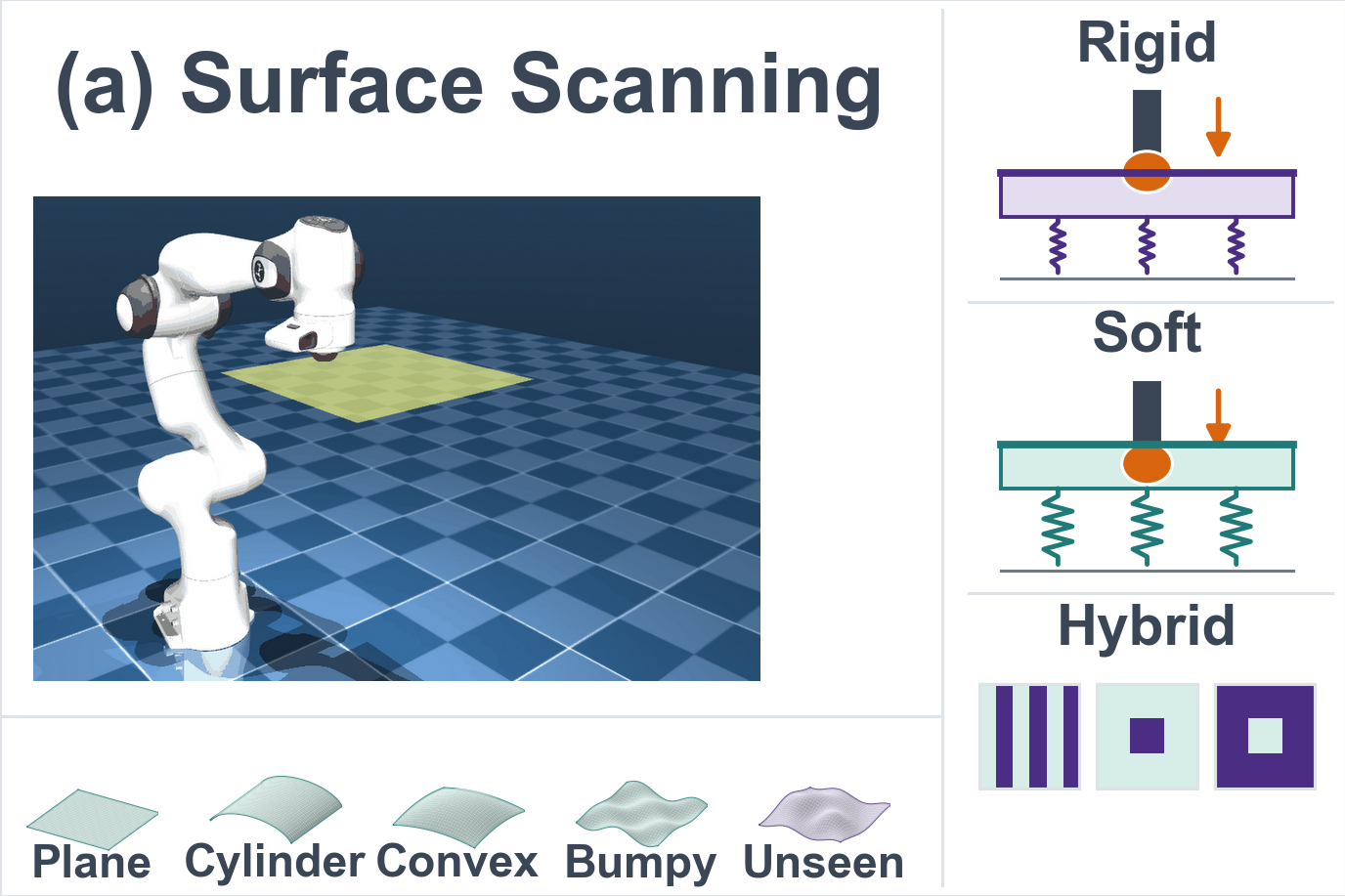}
    \end{minipage}\hfill
    \begin{minipage}[t]{0.32\linewidth}
        \centering
        \includegraphics[width=\linewidth]{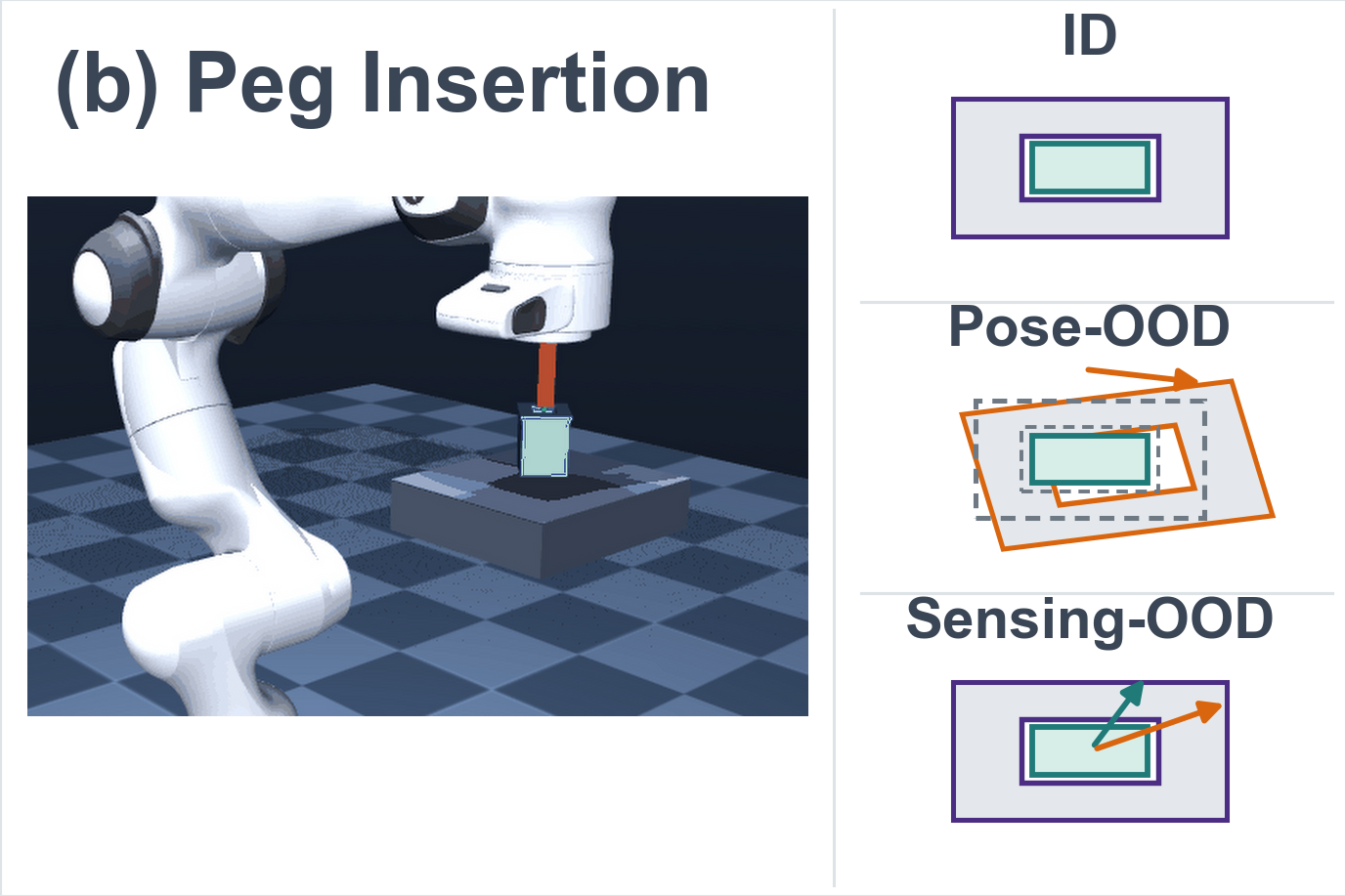}
    \end{minipage}\hfill
    \begin{minipage}[t]{0.32\linewidth}
        \centering
        \includegraphics[width=\linewidth]{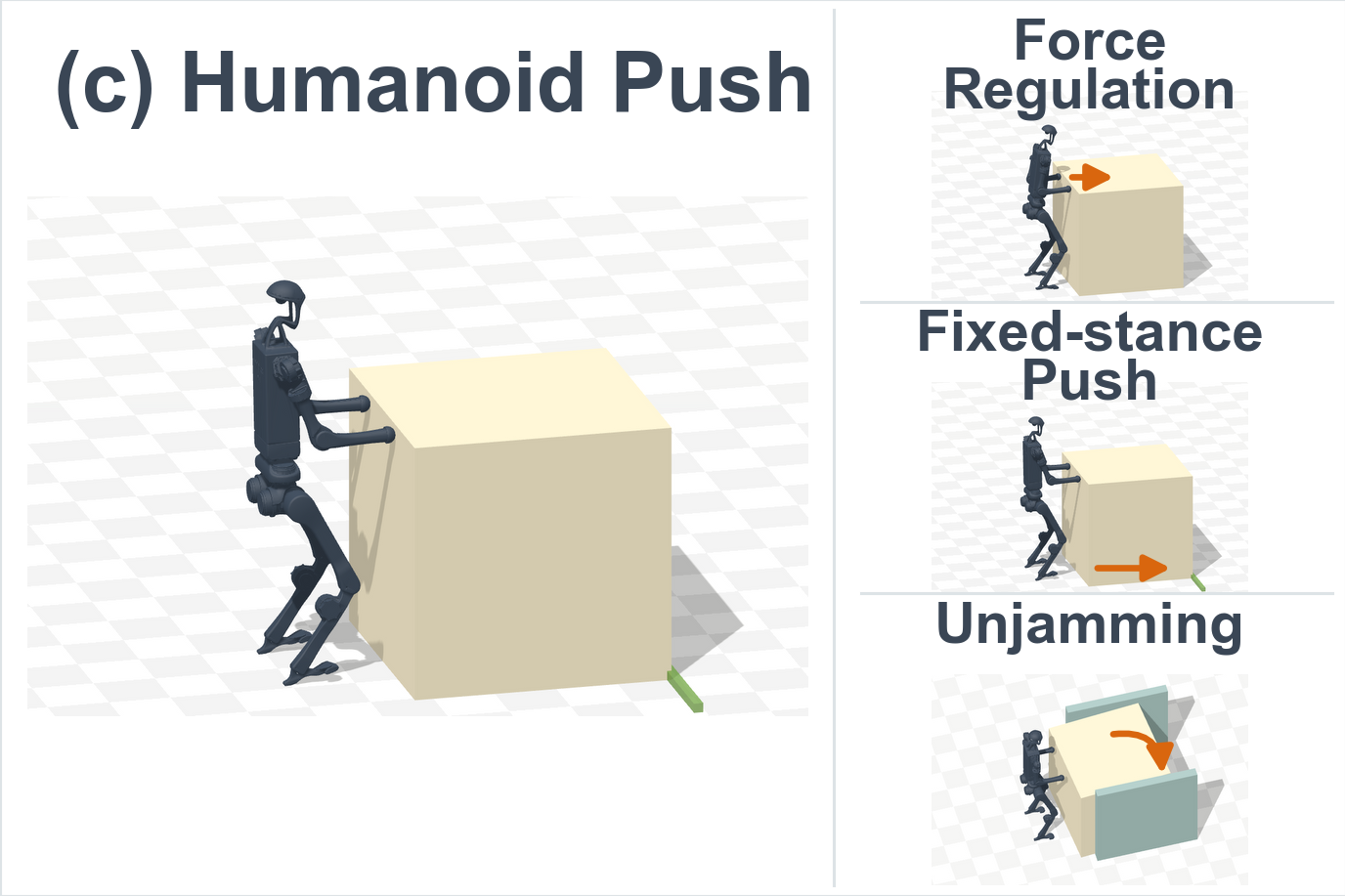}
    \end{minipage}
\begin{minipage}{\linewidth}
\centering
\fontsize{7}{8.5}\selectfont
\setlength{\tabcolsep}{1.6pt}
\renewcommand{\arraystretch}{1.0}
\newcommand{\mgaResult}[2]{#1{\fontsize{3.5}{6}\selectfont$\pm$#2}}
\begin{tabular*}{\textwidth}{@{\extracolsep{\fill}}c|llccccc|>{\columncolor{MGATeal!12}}ccc@{}}
\toprule
\multicolumn{2}{c}{Environment} & Metric
& ISSA & ATACOM & MPPI & DIAL & \shortstack{PegasusFlow} & \multicolumn{1}{c}{\textbf{\MGA{}}} & \shortstack{w/o RL prior} & w/o LRC \\
\specialrule{0.7pt}{3pt}{3pt}
\multirow{8}{*}[-7.2pt]{\rotatebox[origin=c]{90}{\textbf{Surface Scanning}}}
& \multirow{2}{*}{Hard}
& $\mathbf{S}\uparrow$
& \mgaResult{5.0}{10.0} & \mgaResult{25.0}{0.0} & \mgaResult{22.5}{20.8} & \mgaResult{62.5}{20.2}
& \mgaResult{5.0}{10.0} & \mgaResult{\textbf{100.0}}{0.0} & \mgaResult{\textbf{100.0}}{0.0} & \mgaResult{\underline{67.5}}{19.5} \\
&& $\mathbf{F}\downarrow$
& \mgaResult{0.596}{0.050} & \mgaResult{0.533}{0.015} & \mgaResult{1.029}{0.094} & \mgaResult{0.848}{0.122}
& \mgaResult{1.301}{0.237} & \mgaResult{\underline{0.498}}{0.037} & \mgaResult{\textbf{0.486}}{0.017} & \mgaResult{0.521}{0.039} \\
\cmidrule(lr){2-11}
& \multirow{2}{*}{Soft}
& $\mathbf{S}\uparrow$
& \mgaResult{2.5}{7.5} & \mgaResult{50.0}{0.0} & \mgaResult{20.0}{15.0} & \mgaResult{\underline{77.5}}{17.5}
& \mgaResult{30.0}{24.5} & \mgaResult{\textbf{100.0}}{0.0} & \mgaResult{\textbf{100.0}}{0.0} & \mgaResult{70.0}{15.0} \\
&& $\mathbf{F}\downarrow$
& \mgaResult{0.596}{0.051} & \mgaResult{\underline{0.470}}{0.011} & \mgaResult{0.973}{0.059} & \mgaResult{0.762}{0.098}
& \mgaResult{1.094}{0.202} & \mgaResult{\textbf{0.466}}{0.024} & \mgaResult{0.488}{0.031} & \mgaResult{0.498}{0.017} \\
\cmidrule(lr){2-11}
& \multirow{2}{*}{Hybrid}
& $\mathbf{S}\uparrow$
& \mgaResult{0.0}{0.0} & \mgaResult{66.7}{0.0} & \mgaResult{30.0}{10.0} & \mgaResult{56.7}{21.3}
& \mgaResult{20.0}{16.3} & \mgaResult{\textbf{100.0}}{0.0} & \mgaResult{\underline{96.7}}{10.0} & \mgaResult{56.7}{21.3} \\
&& $\mathbf{F}\downarrow$
& \mgaResult{0.698}{0.037} & \mgaResult{\textbf{0.606}}{0.000} & \mgaResult{0.879}{0.034} & \mgaResult{0.789}{0.031}
& \mgaResult{0.862}{0.032} & \mgaResult{\underline{0.658}}{0.016} & \mgaResult{0.738}{0.016} & \mgaResult{0.702}{0.020} \\
\cmidrule(lr){2-11}
& \multirow{2}{*}{Unseen}
& $\mathbf{S}\uparrow$
& \mgaResult{0.0}{0.0} & \mgaResult{50.0}{0.0} & \mgaResult{45.0}{26.9} & \mgaResult{\underline{65.0}}{32.0}
& \mgaResult{15.0}{22.9} & \mgaResult{50.0}{0.0} & \mgaResult{55.0}{15.0} & \mgaResult{\textbf{75.0}}{25.0} \\
&& $\mathbf{F}\downarrow$
& \mgaResult{0.555}{0.025} & \mgaResult{\textbf{0.417}}{0.016} & \mgaResult{0.827}{0.067} & \mgaResult{0.754}{0.149}
& \mgaResult{1.018}{0.353} & \mgaResult{\underline{0.429}}{0.023} & \mgaResult{0.433}{0.018} & \mgaResult{0.502}{0.030} \\
\specialrule{0.7pt}{3pt}{3pt}
\multirow{6}{*}[-4.8pt]{\rotatebox[origin=c]{90}{\textbf{Peg Insertion}}}
& \multirow{2}{*}{ID}
& $\mathbf{S}\uparrow$
& \mgaResult{0.0}{0.0} & \mgaResult{0.0}{0.0} & \mgaResult{60.0}{49.0} & \mgaResult{70.0}{45.8}
& \mgaResult{60.0}{49.0} & \mgaResult{\textbf{100.0}}{0.0} & \mgaResult{\underline{90.0}}{30.0} & \mgaResult{50.0}{50.0} \\
&& $\mathbf{F}\downarrow$
& \mgaResult{0.723}{0.102} & \mgaResult{1.079}{0.000} & \mgaResult{\textbf{0.162}}{0.221} & \mgaResult{0.620}{0.186}
& \mgaResult{0.670}{0.141} & \mgaResult{0.775}{0.010} & \mgaResult{0.733}{0.093} & \mgaResult{\underline{0.449}}{0.289} \\
\cmidrule(lr){2-11}
& \multirow{2}{*}{Pose-OOD}
& $\mathbf{S}\uparrow$
& \mgaResult{0.0}{0.0} & \mgaResult{\underline{50.0}}{50.0} & \mgaResult{30.0}{45.8} & \mgaResult{40.0}{49.0}
& \mgaResult{\textbf{80.0}}{40.0} & \mgaResult{\textbf{80.0}}{40.0} & \mgaResult{\underline{50.0}}{50.0} & \mgaResult{10.0}{30.0} \\
&& $\mathbf{F}\downarrow$
& \mgaResult{1.203}{0.347} & \mgaResult{0.916}{0.053} & \mgaResult{\textbf{0.366}}{0.248} & \mgaResult{0.740}{0.221}
& \mgaResult{0.630}{0.227} & \mgaResult{0.893}{0.093} & \mgaResult{0.769}{0.156} & \mgaResult{\underline{0.477}}{0.365} \\
\cmidrule(lr){2-11}
& \multirow{2}{*}{Sensing-OOD}
& $\mathbf{S}\uparrow$
& \mgaResult{0.0}{0.0} & \mgaResult{0.0}{0.0} & \mgaResult{80.0}{40.0} & \mgaResult{60.0}{49.0}
& \mgaResult{\underline{90.0}}{30.0} & \mgaResult{\textbf{100.0}}{0.0} & \mgaResult{\textbf{100.0}}{0.0} & \mgaResult{30.0}{45.8} \\
&& $\mathbf{F}\downarrow$
& \mgaResult{0.660}{0.053} & \mgaResult{0.969}{0.000} & \mgaResult{\textbf{0.204}}{0.273} & \mgaResult{0.601}{0.228}
& \mgaResult{0.667}{0.089} & \mgaResult{0.774}{0.011} & \mgaResult{0.775}{0.041} & \mgaResult{\underline{0.435}}{0.300} \\
\specialrule{0.7pt}{3pt}{3pt}
\multirow{6}{*}[-4.5pt]{\rotatebox[origin=c]{90}{\textbf{Humanoid Push}}}
& \multirow{2}{*}{Force Regulation}
& $\mathbf{S}\uparrow$
& \mgaResult{\textbf{100.0}}{0.0} & \mgaResult{\textbf{100.0}}{0.0} & \mgaResult{\textbf{100.0}}{0.0} & \mgaResult{\underline{90.0}}{30.0}
& \mgaResult{70.0}{45.8} & \mgaResult{\textbf{100.0}}{0.0} & \mgaResult{\textbf{100.0}}{0.0} & \mgaResult{\underline{90.0}}{30.0} \\
&& $\mathbf{F}\downarrow$
& \mgaResult{0.271}{0.000} & \mgaResult{0.270}{0.000} & \mgaResult{0.288}{0.004} & \mgaResult{0.357}{0.028}
& \mgaResult{0.335}{0.016} & \mgaResult{\textbf{0.262}}{0.005} & \mgaResult{0.280}{0.000} & \mgaResult{\underline{0.263}}{0.004} \\
\cmidrule(lr){2-11}
& \multirow{2}{*}{Fixed-stance Push}
& $\mathbf{S}\uparrow$
& \mgaResult{0.0}{0.0} & \mgaResult{0.0}{0.0} & \mgaResult{\textbf{100.0}}{0.0} & \mgaResult{\textbf{100.0}}{0.0}
& \mgaResult{\textbf{100.0}}{0.0} & \mgaResult{\textbf{100.0}}{0.0} & \mgaResult{\textbf{100.0}}{0.0} & \mgaResult{\textbf{100.0}}{0.0} \\
&& $\mathbf{F}\downarrow$
& \mgaResult{0.972}{0.000} & \mgaResult{1.057}{0.000} & \mgaResult{0.557}{0.019} & \mgaResult{0.550}{0.022}
& \mgaResult{0.559}{0.016} & \mgaResult{\underline{0.503}}{0.038} & \mgaResult{\textbf{0.492}}{0.005} & \mgaResult{0.538}{0.010} \\
\cmidrule(lr){2-11}
& \multirow{2}{*}{Unjamming}
& $\mathbf{S}\uparrow$
& \mgaResult{0.0}{0.0} & \mgaResult{0.0}{0.0} & \mgaResult{\textbf{100.0}}{0.0} & \mgaResult{\textbf{100.0}}{0.0}
& \mgaResult{\textbf{100.0}}{0.0} & \mgaResult{\textbf{100.0}}{0.0} & \mgaResult{\underline{80.0}}{40.0} & \mgaResult{30.0}{45.8} \\
&& $\mathbf{F}\downarrow$
& \mgaResult{1.458}{0.159} & \mgaResult{1.601}{0.000} & \mgaResult{\textbf{0.798}}{0.006} & \mgaResult{\underline{0.805}}{0.008}
& \mgaResult{0.807}{0.012} & \mgaResult{0.845}{0.026} & \mgaResult{0.742}{0.005}$^{\dagger}$ & \mgaResult{0.862}{0.014}$^{\ddagger}$ \\
\bottomrule
\end{tabular*}
\vspace{-0.1cm}
\captionof{table}{Quantitative results of safe-success rate ($\mathbf{S}$) ($\%$), and  normalized contact-load tail risk ($\mathbf{F}$) ($\operatorname{nCVaR}_{95}$) across three contact-rich tasks. Table entries report mean and standard deviation over 10 seeds. Bold and underlining indicate the best and second-best displayed values, including ties.}
\label{tab:contact_main}
\label{tab:peg_insert_main}
\vspace{-0.5cm}
\end{minipage}

\end{figure}


\vspace{-0.2cm}
\paragraph{Lemma 2 (Regularized Geometric Refinement).}
For any clean displacement $d$, let $v=\mathsf M_kd$ and
$H_k=A_kG_k^{-1}A_k^\top$. Then
{%
\setlength{\abovedisplayskip}{2.5pt}
\setlength{\belowdisplayskip}{2.5pt}
\setlength{\abovedisplayshortskip}{1.5pt}
\setlength{\belowdisplayshortskip}{1.5pt}
\begin{equation}
v=\arg\min_z
\left\{
\frac{1}{2}\lVert z-d\rVert_{G_k}^{2}
+\frac{1}{2\lambda_k}\lVert A_kz\rVert_2^{2}
\right\},
\quad
A_kv
=\lambda_k(\lambda_k I+H_k)^{-1}A_kd.
\label{eq:proximal_realization_refinement}
\end{equation}
}%
The minimizer is unique, $\lVert v\rVert_{G_k}\leq\lVert d\rVert_{G_k}$,
and the $G_k$-orthogonal component in $\ker A_k$ is preserved. Every nonzero
encoded normal response is strictly attenuated. Equation~\eqref{eq:ddpm_control_update}
transports this shaped displacement directly into the next clean iterate.
\vspace{-0.3cm}
\paragraph{Theorem 1 (Local Residual Control).}
Suppose $C^{\rm real}(U)=\mathsf J U-b$ on a fixed smooth affine branch containing
$\check U_k$, $V_k=\check U_k+\mathsf M_kd_k$, and the corrected sequence $V_k^+$.
Let $\Pi=\mathsf J\mathsf J^\dagger$ and
$E(U)=\lVert\Pi C^{\rm real}(U)\rVert_2$. For
$0<\eta\leq1$, $\varepsilon>0$, and $\operatorname{rank}(\mathsf J)>0$,
Eq.~\eqref{eq:local_residual_correction} satisfies
{%
\setlength{\abovedisplayskip}{2.5pt}
\setlength{\belowdisplayskip}{2.5pt}
\setlength{\abovedisplayshortskip}{1.5pt}
\setlength{\belowdisplayshortskip}{1.5pt}
\begin{equation}
E(V_k^+)
\leq\rho\left\lVert\Pi C^{\rm real}(\check U_k)+\mathsf J\mathsf M_kd_k\right\rVert_2
=\rho E(V_k),
\quad
\rho=1-\eta\frac{\sigma_{\min,+}^2(\mathsf J)}
{\sigma_{\min,+}^2(\mathsf J)+\varepsilon}<1,
\label{eq:local_residual_control}
\end{equation}
}%
where $\sigma_{\min,+}$ is the smallest positive singular value.
The uncorrectable component remains $-(I-\Pi)b$. It vanishes when
$b\in\operatorname{range}(\mathsf J)$. Lemma~2 controls the encoded response of
the proposed displacement, whereas Theorem~1 reduces the removable chart
residual after transport (Appendix~\appref{app:proofs}).

\begin{figure}[!t]
    \centering
    \edef\mgaMechanismHeight{\the\dimexpr0.5226\linewidth\relax}
    \newcommand{\mgaMechanismFrame}[1]{%
        \begin{tikzpicture}
            \node[draw=MGATeal!50, fill=white, line width=0.5pt,
                  rounded corners=3pt, inner sep=0pt, outer sep=0pt,
                  minimum width=\linewidth, minimum height=\mgaMechanismHeight]
                (frame) {};
            \node[anchor=north, inner sep=0pt, outer sep=0pt, yshift=-3pt]
                at (frame.north)
                {#1};
        \end{tikzpicture}%
    }
    \begin{minipage}[t]{0.63\linewidth}
        \vspace{0pt}
        \centering
        \mgaMechanismFrame{\includegraphics[width=\dimexpr\linewidth-6pt\relax,
                                          height=\dimexpr\mgaMechanismHeight-6pt\relax,
                                          keepaspectratio]{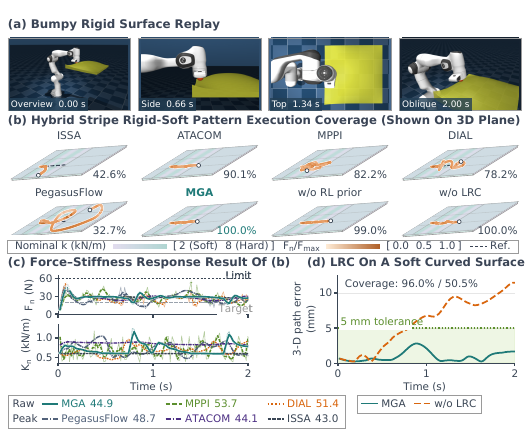}}
    \end{minipage}\hfill
    \begin{minipage}[t]{0.35\linewidth}
        \vspace{0pt}
        \centering
        \mgaMechanismFrame{\includegraphics[width=\dimexpr\linewidth-6pt\relax,
                                          height=\dimexpr\mgaMechanismHeight-6pt\relax,
                                          keepaspectratio]{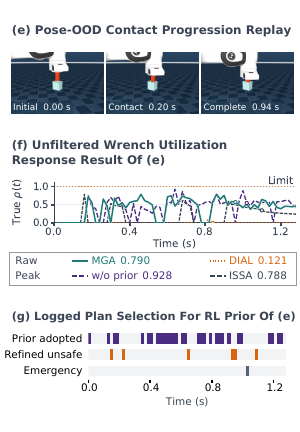}}
    \end{minipage}
    \vspace{-0.15cm}
    \caption{Left: Scanning panels show a (a) bumpy rigid surface replay, (b) hybrid stripe rigid-soft pattern execution coverage and their
    (c) force--stiffness response, and (d) compares \MGA{} and w/o LRC on a soft curved surface.
    Right panels show Pose-OOD Peg Insertion (e) contact progression replay, (f) unfiltered wrench utilization, and (g) logged
    plan selection for RL prior.}
    \label{fig:surface_peg_mechanisms}
    \vspace{-0.4cm}
\end{figure}

\vspace{-0.2cm}
\section{Experimental Analysis}
\label{sec:experiments}
\vspace{-0.2cm}

We evaluate our \MGA{} and its ablation variants through three questions: (\textcolor{MGAGreen}{\textbf{RQ1:}}) Can \MGA{} maintain contact-valid progress while controlling contact force and surface deformation under contact-response mismatch?  (\textcolor{MGAGreen}{\textbf{RQ2:}}) How do learned policy priors complement model-based refinement in achieving success under contact and sensing mismatch?  (\textcolor{MGAGreen}{\textbf{RQ3:}}) Can \MGA{} extend from fixed-base manipulation to whole-body pushing, where balance, contact, and object motion are coupled?

\vspace{-0.2cm}
\paragraph{Experimental Setup.}
We evaluate \textbf{\MGA{}} on three tasks. (\textcolor{MGAGreen}{\textbf{1}}) In Surface Scanning, a robot arm follows a surface path while regulating contact force across rigid and soft surfaces, unseen geometric shapes, and hybrid hard--soft regions. (\textcolor{MGAGreen}{\textbf{2}}) In Peg Insertion, a robot arm inserts a rectangular peg into a fixed socket under nominal conditions (ID), perturbed socket poses (Pose-OOD), and biased force/torque sensing (Sensing-OOD). (\textcolor{MGAGreen}{\textbf{3}}) In Humanoid Push, a Unitree H1 humanoid coordinates balance, hand contact, and box motion during $15$-N force regulation, nominal fixed-stance pushing to the line (green), and unjamming the box with yaw correction. The environments are shown in Fig.~\visualref{fig:environments}. Environment details appear in Appendix~\appref{app:env_setup}.

\vspace{-0.2cm}
\paragraph{Baselines.}
We compare \textbf{\MGA{}} with five baselines. (\textcolor{MGAGreen}{\textbf{1}}) ISSA~\citep{zhao2022modelfree} filters learned policy actions through an Implicit Safe-Set. (\textcolor{MGAGreen}{\textbf{2}}) ATACOM~\citep{liu2022robot} constructs learned actions on an augmented constraint manifold. (\textcolor{MGAGreen}{\textbf{3}}) MPPI~\citep{williams2018information} uses reward-weighted trajectory sampling. (\textcolor{MGAGreen}{\textbf{4}}) DIAL~\citep{xue2025full} performs diffusion-style MPC annealing. (\textcolor{MGAGreen}{\textbf{5}}) PegasusFlow~\citep{ye2025pegasusflow} extends to rolling denoising within the annealing. Two ablations complement these baselines: \textbf{w/o RL prior} removes learned proposals, and \textbf{w/o local residual correction (LRC)} removes LRC. Details appear in Appendix~\appref{app:baseline_details}.

\subsection{Experimental Results}
\label{sec:experimental_results}

\begin{wrapfigure}{R}{0.45\linewidth}
  \centering
  \vspace{-0.35cm}
  \includegraphics[width=\linewidth]{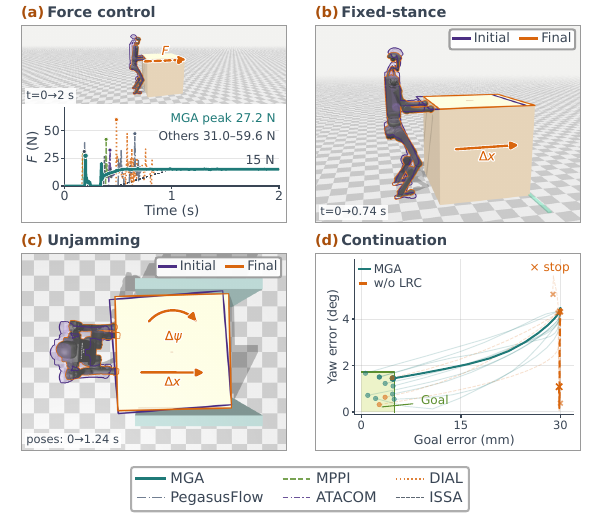}
  \vspace{-0.8cm}
  \caption{Panels (a--c) show force regulation, pushing, and unjamming. Panel (d)
  shows 10 paired position--yaw trajectories.
  Crosses mark safety-triggered stops. Force transients are retained in (a).}
  \label{fig:humanoid_mechanisms}
  \vspace{-0.5cm}
\end{wrapfigure}

\paragraph{Performance across Contact-Rich Tasks.}
As shown in Table~\visualref{tab:contact_main}, on rigid, soft and hybrid surfaces, \MGA{} achieves $100\%$ safe success with lower force tails than MPPI, DIAL, and PegasusFlow, indicating that its coverage is not obtained by pushing harder through contact. In Peg Insertion, \MGA{} achieves $100\%$, $80\%$, and $100\%$ safe success under ID, Pose-OOD, and Sensing-OOD, respectively. It completes insertion in all $30$ trials, but two Pose-OOD trials exceed the force limit. Its higher safe success relative to the tested action-level methods supports the practical value of planning and correcting contact sequences. In Humanoid Push, \MGA{} completes all three tasks safely and lowers tail loads relative to the sampling-based planners during force regulation and fixed-stance pushing. Together, these results support contact-conditioned sequence refinement for preserving task progress while managing interaction loads across the robots and contact regimes.

\begin{figure}[!t]
    \centering
    \includegraphics[width=\linewidth]{figures/exp/real_experiment.jpg}
    \caption{Left shows the hardware setup. Middle shows table, foam, and phantom scanning from top to bottom, with force references of $3$, $0.8$, and $3$~N (Actual: $2.99\pm0.01$, $0.82\pm0.02$, $2.94\pm0.06$~N, respectively). Right plots lateral offset $e$ (mm) against base frame height $z$ (mm) during insertion.}
    \label{fig:real}
    \vspace{-0.4cm}
\end{figure}


\vspace{-0.2cm}
\subsection{Discussion}
\label{sec:experimental_discussion}
\vspace{-0.2cm}

\paragraph{Can \MGA{} Preserve Contact-Valid Progress? (\textcolor{MGAGreen}{RQ1})}
In Fig.~\visualref{fig:surface_peg_mechanisms}(b--c), \MGA{} covers the entire hybrid path while adapting its normal stiffness and keeping contact force below the limit. MPPI, DIAL, and PegasusFlow exhibit larger force peaks but lower coverage, showing that stronger interaction does not necessarily yield useful progress. On the curved soft surface in panel~(d), removing LRC causes the executed path to exceed the $5$-mm tolerance and reduces coverage from $96.0\%$ to $50.5\%$, whereas \MGA{} remains within the tolerance. These matched executions illustrate why geometric shaping alone is insufficient: correcting residuals left by finite updates helps preserve contact-valid motion under changing surface response.

\vspace{-0.2cm}
\paragraph{How Do Learned Priors Complement Model-Based Refinement? (\textcolor{MGAGreen}{RQ2})}
RL-based methods acquire useful insertion behavior, but task competence alone does not ensure safe execution: ISSA and ATACOM complete $29/30$ and $30/30$ trials, yet only $0/30$ and $5/30$ are safe (Fig.~\visualref{fig:app_peg_outcomes}). \MGA{} instead reassesses learned horizons separately and compares them with geometrically refined Gaussian sequences under the additive protocol. Across all three Peg Insertion settings, this combination yields $28/30$ safe completions, versus $17/30$ each for MPPI and DIAL and $23/30$ for PegasusFlow (Table~\visualref{tab:contact_main}). Removing the prior reduces \MGA{} to $24/30$, supporting the value of additional learned candidates under this protocol. Together, these comparisons support combining learned behavioral structure with model-based reassessment and sequence refinement.

\vspace{-0.2cm}
\paragraph{Can \MGA{} Extend to Whole-Body Pushing? (\textcolor{MGAGreen}{RQ3})}
\MGA{} safely completes all three humanoid tasks (Table~\visualref{tab:contact_main}), supporting extension beyond fixed-base manipulation. Fig.~\visualref{fig:humanoid_mechanisms}(d) shows all ten \MGA{} trajectories reaching the position--yaw goal region, whereas w/o LRC completes only three, with seven trials ending when no candidate passes revalidation. Thus, contact-conditioned sequence correction sustains useful continuations under coupled balance, hand contact, and object motion, while retaining the same execution checks.

\subsection{Real-world surface scanning and insertion}

We evaluate the model-based component of \MGA{} on a 6-DoF Dobot Nova~5 with wrist Force/Torque (F/T) sensing (Fig.~\visualref{fig:real}). The robot scans a rigid table, compliant foam, and a curved anatomical phantom while regulating normal force. Foam requires repeated contact recovery, whereas the phantom requires adapting tool height and orientation along the surface. An additional hole-insertion experiment requires alignment before entry and monitored advancement. These observations support its practical use across different contact conditions. Details appear in Appendix~\appref{app:real_experiments}.

\vspace{-0.2cm}
\section{Conclusion}
\label{sec:conclusion}
\vspace{-0.2cm}
We presented \MGA{}, an online motion--impedance optimization method that combines RL policies as structured proposals, model-based reassessment, and feedback-conditioned geometric correction. Rollout weighting identifies useful candidate contributions, geometric shaping adjusts the resulting update, and LRC addresses remaining task residuals. The experiments support the complete refinement method, with learned proposals improving insertion under socket-pose shifts and LRC supporting safe completion across all three insertion conditions. Future work will combine richer proposals from perception models with task-level planning to enable longer-horizon force-control tasks and extend \MGA{} to coordinated multi-robot complex physical interaction settings.

\section{AI Use Disclosure}
\label{sec:ai_use_disclosure}

Large Language Models (LLMs) were employed during the preparation of this manuscript to support both the research and writing processes. Specifically, we used LLMs to assist with research ideation and aspects of research execution, including discussions of potential research directions and support for parts of the research workflow. We also used LLMs to help draft portions of the manuscript and to improve the clarity, readability, grammar, and overall presentation of the text.

The use of LLMs was primarily supportive rather than autonomous. Research ideas, methodological decisions, analyses, and final interpretations were reviewed and determined by the authors. Any LLM-assisted research suggestions or drafted material were critically examined and revised where necessary, and the scientific content and experimental results presented in the paper were independently verified by the authors.

The authors take full responsibility for the final manuscript and all of its contents, including materials produced with the assistance of LLMs. All AI-assisted content was reviewed by the authors to ensure its accuracy, consistency with the actual research, and compliance with applicable ethical and academic standards.
\bibliography{iclr_conference}
\bibliographystyle{iclr_conference}

\newpage
\appendix
\setlength{\floatsep}{8pt plus 2pt minus 2pt}
\setlength{\textfloatsep}{10pt plus 2pt minus 2pt}
\setlength{\intextsep}{7pt plus 2pt minus 2pt}
\renewcommand{\floatpagefraction}{0.85}
\makeatletter
\setlength{\@fptop}{0pt}
\setlength{\@fpsep}{10pt plus 2pt minus 2pt}
\setlength{\@fpbot}{0pt plus 1fil}
\makeatother

\section*{Appendix Contents}

\begingroup
\small
\hypersetup{linkcolor=MGAAppendixInk}
\setcounter{tocdepth}{2}
\makeatletter
\@dottedtocline{1}{0em}{2.4em}{\hyperref[app:related_works]{\ref*{app:related_works}\quad Related Works}}{\pageref*{app:related_works}}
\@dottedtocline{1}{0em}{2.4em}{\hyperref[app:proofs]{\ref*{app:proofs}\quad Theoretical Proofs}}{\pageref*{app:proofs}}
\@dottedtocline{2}{1.6em}{3.2em}{\hyperref[app:theory_assumptions]{\ref*{app:theory_assumptions}\quad Assumptions}}{\pageref*{app:theory_assumptions}}
\@dottedtocline{2}{1.6em}{3.2em}{\hyperref[app:bank_reweighting]{\ref*{app:bank_reweighting}\quad Model-Based Reweighting}}{\pageref*{app:bank_reweighting}}
\@dottedtocline{2}{1.6em}{3.2em}{\hyperref[app:refinement_properties]{\ref*{app:refinement_properties}\quad Properties of Realization-Aware Refinement}}{\pageref*{app:refinement_properties}}
\@dottedtocline{2}{1.6em}{3.2em}{\hyperref[app:residual_control]{\ref*{app:residual_control}\quad Local Residual Control}}{\pageref*{app:residual_control}}
\@dottedtocline{1}{0em}{2.4em}{\hyperref[app:env_setup]{\ref*{app:env_setup}\quad Environmental Setup}}{\pageref*{app:env_setup}}
\@dottedtocline{2}{1.6em}{3.2em}{\hyperref[app:sim_environments]{\ref*{app:sim_environments}\quad Simulation Environments}}{\pageref*{app:sim_environments}}
\@dottedtocline{2}{1.6em}{3.2em}{\hyperref[app:real_environments]{\ref*{app:real_environments}\quad Real-World Deployment Environments}}{\pageref*{app:real_environments}}
\@dottedtocline{1}{0em}{2.4em}{\hyperref[app:baseline_details]{\ref*{app:baseline_details}\quad Baseline Details}}{\pageref*{app:baseline_details}}
\@dottedtocline{1}{0em}{2.4em}{\hyperref[app:real_experiments]{\ref*{app:real_experiments}\quad Real-World Experimental Protocol and Additional Results}}{\pageref*{app:real_experiments}}
\@dottedtocline{2}{1.6em}{3.2em}{\hyperref[app:real_platform]{\ref*{app:real_platform}\quad Deployment Protocol}}{\pageref*{app:real_platform}}
\@dottedtocline{2}{1.6em}{3.2em}{\hyperref[app:real_surface_results]{\ref*{app:real_surface_results}\quad Surface-Scanning Results}}{\pageref*{app:real_surface_results}}
\@dottedtocline{2}{1.6em}{3.2em}{\hyperref[app:real_insert_protocol]{\ref*{app:real_insert_protocol}\quad Hole-Alignment and Insertion Results}}{\pageref*{app:real_insert_protocol}}
\@dottedtocline{1}{0em}{2.4em}{\hyperref[app:realization_maps]{\ref*{app:realization_maps}\quad Task-Specific Realization Maps}}{\pageref*{app:realization_maps}}
\@dottedtocline{1}{0em}{2.4em}{\hyperref[app:additional_results]{\ref*{app:additional_results}\quad Additional Results}}{\pageref*{app:additional_results}}
\@dottedtocline{2}{1.6em}{3.2em}{\hyperref[app:additional_scanning]{\ref*{app:additional_scanning}\quad Surface Scanning}}{\pageref*{app:additional_scanning}}
\@dottedtocline{2}{1.6em}{3.2em}{\hyperref[app:additional_peg_insert]{\ref*{app:additional_peg_insert}\quad Peg Insertion}}{\pageref*{app:additional_peg_insert}}
\@dottedtocline{2}{1.6em}{3.2em}{\hyperref[app:additional_humanoid]{\ref*{app:additional_humanoid}\quad Humanoid Push}}{\pageref*{app:additional_humanoid}}
\@dottedtocline{1}{0em}{2.4em}{\hyperref[app:proposal_implementation]{\ref*{app:proposal_implementation}\quad Proposal and Refinement Details}}{\pageref*{app:proposal_implementation}}
\@dottedtocline{1}{0em}{2.4em}{\hyperref[app:hyperparameters]{\ref*{app:hyperparameters}\quad Hyperparameters}}{\pageref*{app:hyperparameters}}
\@dottedtocline{2}{1.6em}{3.2em}{\hyperref[app:prior_hparams]{\ref*{app:prior_hparams}\quad Learned Proposal Settings}}{\pageref*{app:prior_hparams}}
\@dottedtocline{2}{1.6em}{3.2em}{\hyperref[app:annealing_hparams]{\ref*{app:annealing_hparams}\quad Model-Based Annealing Settings}}{\pageref*{app:annealing_hparams}}
\@dottedtocline{2}{1.6em}{3.2em}{\hyperref[app:refinement_hparams]{\ref*{app:refinement_hparams}\quad Realization-Aware Refinement Settings}}{\pageref*{app:refinement_hparams}}
\@dottedtocline{1}{0em}{2.4em}{\hyperref[app:visualizations]{\ref*{app:visualizations}\quad Visualizations}}{\pageref*{app:visualizations}}
\@dottedtocline{2}{1.6em}{3.2em}{\hyperref[app:vis_scanning]{\ref*{app:vis_scanning}\quad Surface Scanning}}{\pageref*{app:vis_scanning}}
\@dottedtocline{2}{1.6em}{3.2em}{\hyperref[app:vis_peg]{\ref*{app:vis_peg}\quad Peg Insertion}}{\pageref*{app:vis_peg}}
\@dottedtocline{2}{1.6em}{3.2em}{\hyperref[app:vis_humanoid]{\ref*{app:vis_humanoid}\quad Humanoid Push}}{\pageref*{app:vis_humanoid}}
\makeatother
\endgroup

\section{Related Work}
\label{app:related_works}

\paragraph{Safe Reinforcement Learning and Contact-Rich Interaction.}
Safe RL is commonly formulated as a Constrained Markov Decision Process (CMDP), with trust-region, Lagrangian, or Lyapunov updates limiting expected cumulative costs~\citep{achiam2017constrained,chow2019lyapunov}. Because expectation-level constraints may still permit transient violations, safe-exploration methods modify actions through learned safety layers, recovery policies, or implicit safe sets~\citep{dalal2018safe,thananjeyan2021recovery,zhao2022modelfree}. D-ATACOM further learns uncertain long-term safety constraints within constraint-manifold exploration~\citep{gunster2024handling}, whereas FISOR extracts hard-safe offline policies using reachability-based feasibility and diffusion modeling~\citep{zheng2024safe}. For contact-rich tasks, SRL-VIC jointly learns task behavior and variable stiffness with safety-critic and recovery mechanisms~\citep{zhang2024srl}. Together, these methods demonstrate that learned policies contain valuable task- and safety-aware interaction structure. However, they generally execute or safeguard policies at the action level. \MGA{} instead lifts learned behavior into sequence-level proposals whose relevance is reassessed under current contact dynamics.

\paragraph{Model-Based Generative Models for Trajectory Sampling.}
Diffusion planners represent multimodal action or trajectory distributions for offline planning and imitation~\citep{janner2022planning,chi2025diffusion}. Constrained variants impose test-time feasibility through denoising projections, primal--dual guidance, constraint tightening, or augmented Langevin splitting~\citep{romer2024diffusion,christopher2024constrained,zhang2025constrained,blanke2025sal}. Model-Based Diffusion (MBD) instead derives diffusion-style updates directly from objective-weighted dynamics rollouts, enabling training-free trajectory optimization and optional data guidance~\citep{pan2024model}. MDOC and MD-COAS augment this process with barriers, CBF or CFS corrections, and adaptive scheduling~\citep{he2026,he2026motion}, while DIAL-MPC and PegasusFlow extend sampling-based refinement to short-horizon online optimization~\citep{xue2025full,ye2025pegasusflow}. Closest in spirit, JM2D aligns model-free diffusion with model-based modules through interaction-potential joint sampling~\citep{jung2025joint}. These methods establish learned, model-based, and hybrid generative search, but do not use an RL policy as structured proposal source for coupled motion--impedance sequences or refine such candidates through realization-aware geometry.

\paragraph{Constraint Geometry and Manifold-Based Optimization.}
Robot motion generation has long exploited non-Euclidean structure. RMPflow combines local Riemannian motion policies across task spaces, while geometric fabrics shape stable behavior through configurable metrics~\citep{cheng2021rmpflow,van2022geometric}. Constraint-manifold methods such as ATACOM construct safe actions in tangent spaces~\citep{liu2022robot,liu2025safe}, whereas Riemannian score models and flow matching define generative processes intrinsically on manifolds~\citep{de2022riemannian,chen2024flow}. For contact-rich manipulation, geometric impedance learning uses $\mathrm{SE}(3)$-consistent errors and gain scheduling~\citep{seo2023contact}, and log-Euclidean geometry provides unconstrained coordinates for positive-definite stiffness matrices~\citep{arsigny2006log}. CBF-based soft-robot methods map modeled force--deformation limits into invariant pose sets~\citep{dickson2025safe}, while 2GO embeds active-constraint metrics, tangent stochastic filtering, and CFS retraction into model-based generative optimization~\citep{liu2018convex, he20262go}. However, these geometries arise from prescribed constraints, task spaces, or known contact models. \MGA{} uses feedback-conditioned task residuals to shape rollout-weighted sequence updates and correct the residuals left afterward, with response reliability modulating refinement where enabled.

\section{Theoretical Proofs}
\label{app:proofs}

We provide the detailed proofs for Section~\ref{sec:theory}. MBD derives a
model-based clean-moment estimator and establishes small-temperature
concentration of the ideal Gibbs target under sublevel-volume and minimizer
separation assumptions~\citep[Section~4.1 and Appendix~A.2]{pan2024model}.
DIAL similarly relates MPPI to a Gaussian-smoothed score, followed by a
Monte Carlo approximation~\citep[Section~III-A]{xue2025full}. These results do
not establish convergence of our finite, structured proposal bank. Likewise,
the residual recursion in MD-COAS assumes bounds on proposal-induced residual
growth and local correction error~\citep{he2026motion}. Below, we instead derive
the finite-bank weighting identity and the residual response of the specific
normalized update used by \MGA{}.

\subsection{Assumptions}
\label{app:theory_assumptions}

\begin{description}
\item[A1 (Evaluated proposal bank).]
We condition on the $N_k\geq1$ indexed candidates and their finite costs
$j_m=\widetilde J(\widetilde U_{k,m})$. The resulting temperature
$\beta_k>0$ is fixed, including when it is computed from these costs.
Repeated candidates, reused learned horizons, and deterministic anchors are
allowed. Each evaluated slot retains its multiplicity.

\item[A2 (Fixed geometric core).]
Within one reverse step, $A_k$ and the symmetric positive-definite metric
$G_k$ are fixed, and $\lambda_k>0$. No full-rank assumption on $A_k$ is
required.

\item[A3 (Local correction chart).]
For Theorem~1, the feedback-conditioned chart is affine,
$C^{\rm real}(U)=\mathsf J U-b$, on one smooth branch containing
$\check U_k$, $V_k=\check U_k+\mathsf M_kd_k$, and $V_k^+$.
The points lie away from clipping-switch boundaries, and the correction
uses $0<\eta\leq1$ and $\varepsilon>0$.
Rank deficiency and an incompatible target $b$ are allowed. The smooth
nonlinear extension is stated separately below.
\end{description}

\subsection{Model-Based Reweighting}
\label{app:bank_reweighting}

\paragraph{Proof of Lemma 1.}
Write $N=N_k$, $\beta=\beta_k$, $u=\mathbf1/N$, and
$Z=N^{-1}\sum_m\exp(-j_m/\beta)$. The softmax weights satisfy
$\log(\bar w_{k,m}/u_m)=-j_m/\beta-\log Z$. Hence, for every
$w\in\Delta_N$, with the convention $0\log0=0$,
\begin{equation}
\underbrace{\sum_m w_mj_m+\beta D_{\mathrm{KL}}(w\|u)}_{F(w)}
=-\beta\log Z+\beta D_{\mathrm{KL}}(w\|\bar w_k).
\label{eq:bank_gibbs_identity}
\end{equation}
Nonnegativity of KL divergence, with equality only at $w=\bar w_k$,
proves the unique minimizer. Comparing $F(\bar w_k)$ with $F(u)$ and
with a point mass on a minimum-cost candidate gives
\begin{equation}
\min_m j_m\leq\sum_m\bar w_{k,m}j_m
\leq\min\!\left\{\frac1{N_k}\sum_mj_m,
\ \min_mj_m+\beta_k\log N_k\right\}.
\label{eq:bank_cost_bound}
\end{equation}
The lower bound follows because the weights form a convex combination.
\hfill$\square$

This identity is the Gibbs tilt of the indexed empirical measure
$N_k^{-1}\sum_m\delta_{\widetilde U_{k,m}}$, not an importance-corrected
estimator of the continuous target in Eq.~\eqref{eq:soft_target_distribution}.
No independence or proposal density is needed. In particular, the bound does
not extend to $\widetilde J(\widehat{\bar U}_k)$ for a nonconvex objective.

\paragraph{Coverage under a fixed proposal budget.}
The role of the learned prior can be separated from this weighting result.
Consider the initial sampling step with fixed state and proposal maps, and
a measurable useful set $\mathcal E$ of positive thickness. Suppose
independently drawn, mapped Gaussian and learned slots
hit $\mathcal E$ with probabilities $p_{\rm g}$ and $p_\psi$, respectively.
For the fixed anchor set $\mathcal A_k^{\rm anc}$ from Section~\ref{subsec:method_rl_prior},
\begin{equation}
\Pr\!\left((\mathcal Q_k\cup\mathcal A_k^{\rm anc})\cap\mathcal E=\emptyset\right)
=\mathbf1\{\mathcal A_k^{\rm anc}\cap\mathcal E=\emptyset\}
(1-p_{\rm g})^{M_{\rm g}}(1-p_\psi)^{M_\psi}.
\label{eq:prior_bank_coverage}
\end{equation}
This follows by multiplying the independent miss probabilities. At fixed
$M_{\rm g}+M_\psi=M$, replacing Gaussian slots improves or preserves coverage
if $p_\psi\geq p_{\rm g}$. This is a conditional advantage, not a guarantee
for any pretrained policy. The comparison concerns the initial draw with
fixed maps. Cached learned horizons are not independent fresh draws at later
reverse steps. Coverage alone does not order the final refined sequences.

\subsection{Properties of Realization-Aware Refinement}
\label{app:refinement_properties}

\paragraph{Proof of Lemma 2.}
Write $G=G_k$, $A=A_k$, and $\lambda=\lambda_k$. The quadratic objective
is strongly convex, and its first-order condition is
$G(v-d)+\lambda^{-1}A^\top Av=0$. The matrix inversion lemma gives
\begin{equation}
v
=(G+\lambda^{-1}A^\top A)^{-1}Gd
=\left[I-G^{-1}A^\top
(\lambda I+AG^{-1}A^\top)^{-1}A\right]d
=\mathsf M_kd.
\end{equation}
Multiplication by $A$ yields the second identity in
Eq.~\eqref{eq:proximal_realization_refinement}. For $B=AG^{-1/2}$,
\begin{equation}
G^{1/2}\mathsf M_kG^{-1/2}
=I-B^\top(\lambda I+BB^\top)^{-1}B
=(I+\lambda^{-1}B^\top B)^{-1}.
\label{eq:whitened_geometric_filter}
\end{equation}
Its eigenvalues are one on $\ker B$ and
$\lambda/(\lambda+\sigma_i^2)\in(0,1)$ on the right singular directions
with $\sigma_i>0$. Thus $\lVert v\rVert_G\leq\lVert d\rVert_G$, and
the $G$-orthogonal component in $\ker A$ is unchanged. Since
$Ad\in\operatorname{range}(A)=\operatorname{range}(AG^{-1}A^\top)$,
the response identity strictly contracts $Ad$ whenever $Ad\neq0$.
When $A$ is the complete active Jacobian at a regular feasible point,
the limit $\lambda\downarrow0$ recovers the metric tangent map in
Eq.~\eqref{eq:metric_tangent_projection}.
\hfill$\square$

\paragraph{Clean-moment transport.}
Substituting Eq.~\eqref{eq:mb_score_U} into
Eq.~\eqref{eq:ddpm_control_update}, and using
$\bar\alpha_k=\alpha_k\bar\alpha_{k-1}$, gives
\begin{equation*}
U_{k-1}^{\rm raw}
=\frac{1}{\sqrt{\alpha_k}}
\left[U_k-U_k+\sqrt{\bar\alpha_k}\bar U_{\mathcal M,k}\right]
=\sqrt{\bar\alpha_{k-1}}\bar U_{\mathcal M,k}.
\end{equation*}
Consequently, the transported clean sequence is
$V_k=\check U_k+\mathsf M_kd_k$, and Lemma~2 gives
\begin{equation}
A_k(V_k-\check U_k)
=\lambda_k(\lambda_k I+H_k)^{-1}A_kd_k,
\quad
\lVert A_k(V_k-\check U_k)\rVert_2\leq\lVert A_kd_k\rVert_2.
\label{eq:normal_contraction_transport}
\end{equation}
The inequality is strict when $A_kd_k\neq0$. This identity concerns the
chosen deterministic transport. Arbitrary reverse transports or additional
reverse noise need not preserve this samplewise bound.

\paragraph{From encoded responses to chart responses.}
Lemma~2 controls $A_kd$, not necessarily the full Jacobian response.
To quantify this distinction, write $A=A_k$, $G=G_k$, $\lambda=\lambda_k$,
and let $\mathsf M_A$, $\mathsf M_{\mathsf J}$ be the same regularized filter
constructed from $A$ and a fixed chart Jacobian $\mathsf J$, respectively.
Set $B=AG^{-1/2}$, $D=\mathsf JG^{-1/2}$, and
$\delta=\lVert B^\top B-D^\top D\rVert_2$. With
$R_B=(I+B^\top B/\lambda)^{-1}$ and
$R_D=(I+D^\top D/\lambda)^{-1}$, the resolvent identity gives
\begin{math}
R_B-R_D=\lambda^{-1}R_B(D^\top D-B^\top B)R_D.
\end{math}
Both resolvents have norm at most one, so
$\lVert(\mathsf M_A-\mathsf M_{\mathsf J})d\rVert_G
\leq(\delta/\lambda)\lVert d\rVert_G$. If $D$ has positive rank,
Lemma~2 applied to $\mathsf J$ and the triangle inequality yield
\begin{equation}
\lVert\mathsf J\mathsf M_A d\rVert_2
\leq
\frac{\lambda}{\lambda+\sigma_{\min,+}^2(D)}\lVert\mathsf Jd\rVert_2
+\frac{\lVert D\rVert_2\delta}{\lambda}\lVert d\rVert_G.
\label{eq:encoded_jacobian_bound}
\end{equation}
The zero-rank case is trivial. This bound makes the required encoding accuracy
explicit. It does not assume that the implemented dictionary has small
$\delta$. Its construction and limitations are described in
Appendix~\appref{app:geometry_implementation}.

\subsection{Local Residual Control}
\label{app:residual_control}

\paragraph{Proof of Theorem 1.}
On the fixed affine branch, abbreviate $C=C^{\rm real}$ and define
$T_\varepsilon:=I-\eta\mathsf J\mathsf J^\top
(\mathsf J\mathsf J^\top+\varepsilon I)^{-1}$. Substitution of
Eq.~\eqref{eq:local_residual_correction} gives the exact identity
\begin{equation}
C(V_k^+)=T_\varepsilon C(V_k)
=T_\varepsilon\left[C(\check U_k)+\mathsf J\mathsf M_kd_k\right].
\label{eq:composed_residual_identity}
\end{equation}
The orthogonal projector $\Pi=\mathsf J\mathsf J^\dagger$ commutes with
$T_\varepsilon$. On each left singular direction of $\mathsf J$ with
singular value $\sigma_i>0$, the latter has eigenvalue
$1-\eta\sigma_i^2/(\sigma_i^2+\varepsilon)\in[0,\rho]$.
Therefore
\begin{math}
\lVert\Pi C(V_k^+)\rVert_2
\leq\rho\lVert\Pi C(V_k)\rVert_2,
\end{math}
and $\Pi\mathsf J=\mathsf J$ proves Eq.~\eqref{eq:local_residual_control}.
On the orthogonal complement, $T_\varepsilon$ is the identity and
$(I-\Pi)C(U)=-(I-\Pi)b=:c_\perp$ is constant. Equivalently,
\begin{equation}
\lVert C(V_k^+)\rVert_2^2-\lVert c_\perp\rVert_2^2
\leq\rho^2\left[\lVert C(V_k)\rVert_2^2-\lVert c_\perp\rVert_2^2\right].
\label{eq:affine_residual_floor}
\end{equation}
If $\mathsf J=0$, the correction is zero and $C(V_k^+)=C(V_k)=-b$.
Thus redundant rows require no full-row-rank assumption, while incompatible
components cannot be removed by this correction.
\hfill$\square$

The task charts in Appendix~\appref{app:realization_maps} are affine after
fixing the replanning context and the clipping branches. Theorem~1 therefore
applies when the update remains on one such branch. It compares $V_k^+$ with
$V_k$, not with $\check U_k$: the proposal displacement may introduce residual
before correction. Neither branch switching nor a sequence of changing
proposal banks is covered by a fixed-chart convergence claim.

\paragraph{Smooth nonlinear charts.}
Let $C=C^{\rm real}$ have an $L$-Lipschitz Jacobian on the segment from
$V_k$ to $V_k^+=V_k+p$, where $p$ is the correction in
Eq.~\eqref{eq:local_residual_correction}, with $0<\eta\leq1$ and
$\varepsilon>0$. With
$\mathsf J=\mathsf J(V_k)$, the integral Taylor formula gives
\begin{equation*}
C(V_k^+)=T_\varepsilon C(V_k)+e,
\quad
e=\int_0^1[\mathsf J(V_k+sp)-\mathsf J(V_k)]p\,ds,
\quad \lVert e\rVert_2\leq\frac L2\lVert p\rVert_2^2.
\end{equation*}
For $\Pi=\mathsf J\mathsf J^\dagger$ and the positive-rank factor $\rho$
defined in Theorem~1, the same orthogonal decomposition yields
\begin{equation}
\lVert C(V_k^+)\rVert_2
\leq\sqrt{\lVert(I-\Pi)C(V_k)\rVert_2^2
+\rho^2\lVert\Pi C(V_k)\rVert_2^2}
+\frac L2\lVert p\rVert_2^2.
\label{eq:nonlinear_residual_bound}
\end{equation}
Thus the linearized reduction persists when it exceeds the second-order
remainder. For example, if $C(V_k)\in\operatorname{range}(\mathsf J)$ and
$L\lVert p\rVert_2^2/2<(1-\rho)\lVert C(V_k)\rVert_2$, the full residual
strictly decreases. Unlike the affine case, $(I-\Pi)C(V_k)$ is only a local
linearization component, not a fixed residual floor for the nonlinear chart.
If $\mathsf J(V_k)=0$, then $p=0$ and the correction leaves $V_k$ unchanged.

\section{Environmental Setup}
\label{app:env_setup}

We provide detailed descriptions of the environments used in this work.
The simulation benchmarks isolate three aspects of contact-rich control:
\textbf{(1)} adaptation of learned insertion behavior to model and sensing
mismatch, \textbf{(2)} coordination of path progress, force, and deformation
under changing surface response, and \textbf{(3)} whole-body balance during
object interaction. The real-world environments examine deployment of the
model-based refinement component under measured contact feedback.

\subsection{Simulation Environments}
\label{app:sim_environments}

\paragraph{PegInsert.}
A Franka Panda inserts a rectangular peg into a fixed socket by coordinating
pose alignment, impedance, and axial force. The peg has a $20\times12$-mm
cross-section and a $50$-mm length. The socket is $40$~mm deep, with
$1.5$-mm nominal clearance on each side. The peg starts $4$~mm above the
socket. We use a $10$-N axial-force reference, force limits of $30$~N axially
and $20$~N laterally, and bending/torsional torque limits of $1.5/1.0$~N\,m.
The three evaluation conditions retain the same task objective but change
the source of mismatch:
\begin{itemize}
\item \textbf{ID:} We use nominal socket geometry and unbiased wrench observations.
\item \textbf{Pose-OOD:} Execution-side socket translations are perturbed
independently by up to $1$~mm in each lateral direction, and orientation
components by up to $0.7^\circ$, while the planner retains the nominal model.
Clearance and friction are unchanged.
\item \textbf{Sensing-OOD:} Execution observations include force bias
$(0.5,-0.5,0)$~N and torque bias $(0.02,-0.02,0.01)$~N\,m, without changing
the physical task geometry.
\end{itemize}
Completion requires at least $32$~mm of insertion, lateral error at most
$1.2$~mm, orientation error at most $0.06$~rad, and instantaneous wrench
compliance for three consecutive control steps. Strict safe success further
requires a complete $64$-step execution with no wrench violation or jam
anywhere in that window, including steps after task completion. Execution
continues through all $64$ steps. Incomplete or aborted runs count as failures.
Safety metrics use the true simulated wrench rather than the biased
observations. This distinction separates successful insertion from unsafe
progress and isolates the role of model-based reassessment of learned behavior.

\paragraph{Surface Scanning.}
A Panda follows a prescribed surface path while regulating normal force and
limiting deformation. We consider five shape families---plane, cylinder,
convex, bumpy, and held-out randomly generated surfaces---under both rigid
and compliant contact. Three additional planar environments vary material
response spatially through alternating hard--soft stripes, a hard center,
or a soft center, giving $13$ conditions in total. Force references are
$20$~N for rigid and hybrid surfaces, $18$~N for the four nominal compliant
shapes, and $15$~N for the unseen compliant surface. The force limit is
$60$~N. The unseen setting also varies friction and compliance.

The main table partitions these conditions into four disjoint groups.
\textbf{Hard} contains rigid plane, cylinder, convex, and bumpy surfaces.
\textbf{Soft} contains their four compliant counterparts.
\textbf{Hybrid} contains the three spatial stiffness maps.
\textbf{Unseen} contains the held-out rigid and compliant surfaces.
Unseen trials are not also counted in Hard or Soft.

Rigid and compliant surfaces use MuJoCo probe--surface contact. The hybrid
environments instead use a spatial elastic-foundation response with hard
and soft stiffnesses of $8000$ and $2000$~N/m, respectively, without an
additional probe collision response. This separates material transitions
from changes in surface shape. We evaluate realized in-contact path progress,
force tracking, contact loss, and deformation, rather than commanded path
progress alone. These environments test whether refinement follows a
physically responsive direction when similar commands produce different
contact motions.

Scanning retains the operational SSR criterion frozen after the two-seed
layout study: contact-valid trajectory coverage must be at least $90\%$,
settled contact loss at most $10\%$, and force-violating samples at most
$1\%$. Coverage uses the complete executed path with a $5$-mm tolerance.
Only contact-loss measurement excludes the first ten acquisition steps.
The force tolerance allows at most one of the $100$ control-step samples
to exceed $60$~N and is not a zero-violation criterion. Tail utilization is
the mean of the five largest actual normal-force samples divided by $60$~N.
It does not measure deformation or force-tracking error.

\paragraph{Humanoid Push.}
A Unitree H1 coordinates hand contact, whole-body balance, and box motion.
The reported comparison contains three conditions. \textbf{Force Regulation}
tracks a $15$-N hand-force reference against an effectively stationary box,
isolating force control from object displacement. \textbf{Fixed-stance Push}
moves a nominal $8$-kg box $0.10$~m from a fixed stance.
\textbf{Unjamming} releases an initially rotated
box from a narrow corridor and advances it $0.03$~m, with a final yaw
tolerance of $0.03$~rad. The $30$-N force step, randomized pushing
conditions, and walking pushes are not part of this comparison.

Force-regulation SSR requires reaching $90\%$ of the force reference, staying within
$15\pm1.5$~N over the terminal $0.20$~s, and avoiding physical-safety violations.
This criterion measures force-regulation success rather than box displacement.
Fixed-stance pushing and unjamming require task completion without physical-safety violations,
including excessive hand force, unintended body--object contact, or
violations of balance and fall limits. Safety checks include the initial
state and all active $4$-ms physics samples. The normalized tail metric
uses physical hand force divided by the $60$-N limit and averages the
largest $\lceil0.05N\rceil$ samples. Post-success padding is excluded.
It does not subsume the separate balance and contact constraints.

\subsection{Real-World Deployment Environments}
\label{app:real_environments}

\begin{figure}[!htbp]
\centering
\includegraphics[width=\linewidth]{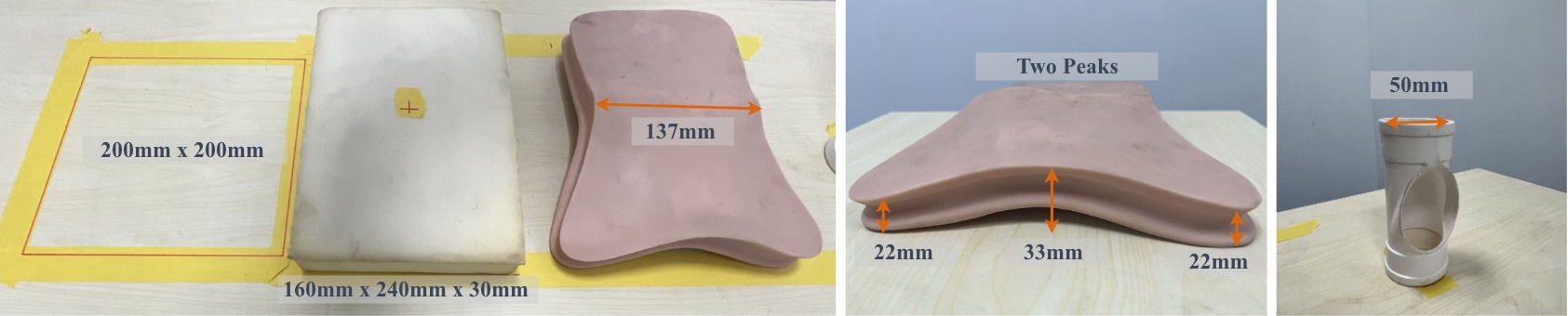}
\caption{Real-world contact fixtures: the rigid table region, compliant foam,
and curved phantom (left), the phantom profile (center), and the pipe used
for clearance insertion (right). Dimensions describe the fixtures rather
than the commanded path lengths.}
\label{fig:app_real_setup}
\end{figure}

The hardware platform is a 6-DoF Dobot Nova~5 with a wrist-mounted 6-axis
force/torque sensor and a probe surrogate having a $40$-mm-diameter contact
head and a $200$-mm nominal length. We use three scanning surfaces and one
insertion fixture to separate material compliance, curved-surface following,
and pre-contact geometric alignment (Fig.~\visualref{fig:app_real_setup}).

\paragraph{Table.}
A rigid planar table provides a reference condition for contact establishment
and steady force tracking. The robot traverses a $100$-mm path with a
$3$-N normal-force reference.

\paragraph{Foam.}
A nearly planar compliant foam block uses the same $100$-mm path with a
lower $0.8$-N force reference. This environment emphasizes contact recovery
and the tradeoff between tangential progress and force regulation.

\paragraph{Phantom.}
A curved anatomical phantom requires a $110$-mm scan at a $3$-N force
reference. The controller estimates the local normal from recent contact
positions and adjusts tool orientation, without a pre-reconstructed surface
mesh or live camera feedback.

\paragraph{Hole Insertion.}
A $50$-mm-bore pipe accommodates the $40$-mm tool envelope, giving a nominal
$5$-mm radial clearance. The robot begins $100$~mm above the rim with an
approximately $40$-mm lateral offset, aligns with the pipe axis, and then
inserts the tool. This is a clearance-insertion task, distinct from the
tight-clearance simulated PegInsert benchmark.

These are bench contact-control experiments with a surrogate tool, not
ultrasound-image acquisition or human-subject experiments. The deployment
protocol and quantitative results are reported in
Appendix~\appref{app:real_experiments}.

\section{Baseline Details}
\label{app:baseline_details}

We compare action-level safety correction and sampling-based trajectory
optimization to separate local constraint handling from model-based horizon
refinement. Table~\visualref{tab:method_attribute}
summarizes the task-adapted implementations and the \MGA{} variants.
The descriptions below explain how each method uses the common contact-control
interface. The entries do not imply global safety guarantees.

\paragraph{ISSA.}
The implicit safe-set algorithm~\citep{zhao2022modelfree} filters the same
nominal PPO action through a derivative-free search. Its safety test queries
one-step transitions to discourage unsafe set crossings or avoid increasing
the safety index when already outside the safe set. It does not optimize a trajectory
horizon. Thus, ISSA tests action-level correction rather than joint proposal
sampling and geometric sequence refinement.

\paragraph{ATACOM.}
ATACOM~\citep{liu2022robot} transforms a learned free action into a direction
on an augmented constraint manifold, using slack variables for inequalities
and a local correction for manifold drift. We adapt this construction to the
common contact-control primitive and train a dedicated PPO policy through
the tangent-action interface. It is not a post-hoc projection of the ordinary
PPO policy. This baseline distinguishes action-level constraint geometry
from feedback-conditioned sequence refinement.

\begin{table*}[tb!]
\small
\newcolumntype{O}{>{\arraybackslash}m{0.35cm}}
\newcolumntype{A}{>{\raggedright\arraybackslash}m{2.05cm}}
\newcolumntype{B}{>{\centering\arraybackslash}m{1.45cm}}
\newcolumntype{C}{>{\centering\arraybackslash}m{1.10cm}}
\newcolumntype{D}{>{\centering\arraybackslash}m{1.25cm}}
\newcolumntype{E}{>{\centering\arraybackslash}m{2.40cm}}
\newcolumntype{F}{>{\centering\arraybackslash}m{1.70cm}}
\newcolumntype{G}{>{\centering\arraybackslash}m{0.80cm}}
\newcolumntype{H}{>{\centering\arraybackslash}m{1.50cm}}
\setlength\tabcolsep{2pt}
\renewcommand{\arraystretch}{1.5}
\centering
\begin{tabular}{O|ABCDEFGH}
\toprule
\multicolumn{2}{c}{\textbf{Method}}
& \textbf{Policy Role}
& \textbf{Model-based}
& \textbf{Horizon Rollout}
& \textbf{Constraint Treatment}
& \textbf{Realization Geometry}
& \textbf{LRC}
& \textbf{Learned Reliability}\\
\midrule
\parbox[t]{3mm}{\multirow{5}{*}{\rotatebox[origin=c]{90}{\textbf{Baselines}}}}
& ISSA & Nominal action & \ding{55} & \ding{55} & Implicit safe-set filter & \ding{55} & \ding{55} & \ding{55}\\
& ATACOM & Tangent action & \ding{55} & \ding{55} & Augmented constraint manifold & \ding{55} & \ding{55} & \ding{55}\\
& MPPI & \ding{55} & \ding{51} & \ding{51} & Reward penalties & \ding{55} & \ding{55} & \ding{55}\\
& DIAL & \ding{55} & \ding{51} & \ding{51} & Reward penalties & \ding{55} & \ding{55} & \ding{55}\\
& PegasusFlow & \ding{55} & \ding{51} & \ding{51} & Reward penalties & \ding{55} & \ding{55} & \ding{55}\\
\midrule
\parbox[t]{3mm}{\multirow{3}{*}{\rotatebox[origin=c]{90}{\textbf{Variants}}}}
& \textbf{\MGA{}} & Structured proposals & \ding{51} & \ding{51} & Feasibility cost + local correction & \ding{51} & \ding{51} & \ding{51}\\
& \MGA{} w/o RL prior & \ding{55} & \ding{51} & \ding{51} & Feasibility cost + local correction & \ding{51} & \ding{51} & \ding{51}\\
& \MGA{} w/o LRC & Structured proposals & \ding{51} & \ding{51} & Feasibility cost without local correction & \ding{51} & \ding{55} & \ding{51}\\
\bottomrule
\end{tabular}
\caption{The table compares the configured simulation baselines and \MGA{} variants.
Entries describe the task-adapted implementations, with \ding{51} and
\ding{55} denoting presence and absence of the indicated mechanism.
Model-based denotes use of an explicit dynamics model to predict and
optimize candidate trajectories. Black-box safety queries and local
constraint geometry do not qualify. Realization geometry
and LRC denote response-conditioned sequence refinement and its
post-transport residual correction, not action-level constraint handling.}
\label{tab:method_attribute}
\end{table*}

\paragraph{MPPI.}
Model Predictive Path Integral control~\citep{williams2018information}
perturbs dense control sequences with Gaussian noise and combines them using
rollout-return weights. The task reward penalizes undesirable contact
behavior, but the update has neither a learned proposal prior nor the
realization-aware geometric filter.

\paragraph{DIAL.}
DIAL~\citep{xue2025full} uses diffusion-style annealing of horizon-node
perturbations with reward-weighted sequence updates. We retain this sampling
mechanism under the common task interface, without \MGA{}'s learned
proposals, additional feasibility cost, or local manifold correction.
It tests whether annealing alone suffices under contact-response mismatch.

\paragraph{PegasusFlow.}
PegasusFlow~\citep{ye2025pegasusflow} performs rolling denoising with adaptive
basis-weighted updates of a spline-parameterized horizon. Like DIAL, it
evaluates candidate controls through task rollouts but does not use the
learned proposal or realization-aware correction of \MGA{}.

\paragraph{No RL Prior (w/o RL prior).}
This ablation removes the proposal policy while retaining the remaining
refinement and execution-selection mechanisms of \MGA{}.

\paragraph{No Local Residual Correction (w/o LRC).}
This ablation disables the correction in Eq.~\eqref{eq:local_residual_correction}
while retaining learned proposals, model-based rollout weighting, geometric
shaping, realization-aware geometry, and execution acceptance. 

\paragraph{\MGA{}.}
Our method uses the policy to supply structured candidate horizons rather
than committing to its action. Under replacement, rollout weighting combines
learned and Gaussian candidates before realization-aware refinement. Under
additive integration, separately evaluated learned horizons compete with the
refined Gaussian sequence through execution acceptance. Comparing with the
no-prior and no-LRC ablations therefore tests the complementary roles of
learned structure, model-based reassessment, and local correction.

\paragraph{Common comparison settings.}
Methods share the task interface, control period, evaluation conditions, and
random seeds. Sampling-based methods use a nominal budget of $64$ stochastic
proposals and the same refinement-iteration settings within each benchmark.
Method-specific anchors, safety queries, and revalidation are retained.
This matches the nominal sampling budget, not the total computational cost.
In all three PegInsert suites and the reported humanoid runs, the prior is
evaluated additively:
one deterministic and eight stochastic policy proposals are evaluated once
per replanning call after Gaussian refinement and revalidation, with $64$
Gaussian samples per refinement iteration. The selected expert can replace
the refined Gaussian sequence if accepted. These expert horizons do not enter
the Gaussian refinement bank or its geometric correction. Its comparison with the no-prior
ablation therefore does not establish an advantage at equal total rollout cost.
The environment reward is shared, while additional feasibility weighting
is part of the \MGA{} variants. In PegInsert, ISSA and \MGA{} use the same
frozen PPO policy. ATACOM uses a separately trained tangent-space
policy with the same two-layer, $16$-unit architecture. Both policies use
$200{,}000$ environment steps of training and select the final-budget
checkpoint after an independent development health gate. Each policy's
checkpoint hash is locked across all three PegInsert suites. No policy is
updated on the evaluation trials.

\section{Real-World Experimental Protocol and Additional Results}
\label{app:real_experiments}

We report the deployment protocol and additional measurements for the
environments in Appendix~\appref{app:real_environments}. These experiments
evaluate the model-based refinement component of \MGA{} with contact
feedback. The hardware configuration does not use a learned proposal prior.

\subsection{Deployment Protocol}
\label{app:real_platform}
\label{app:real_surface_protocol}

The planner uses a $0.4$-s horizon, $48$ candidate sequences, and three
initial or two subsequent refinement iterations. Planning and position-command
execution have nominal rates of $10$ and $20$~Hz, respectively. A bounded
normal-force feedback controller converts the selected force reference into
position corrections, while tangential progress depends on the measured
contact response. This is a position-based realization of the \MGA{}
deployment component, not direct execution of a full Cartesian stiffness
matrix. The observed behavior reflects the planner and feedback controller
together.

For surface scanning, each trial begins with unloaded force/torque zeroing
and a slow approach to the taught path. After contact is established and
force settles near its reference, the robot traverses the path at a nominal
commanded speed of $5$~mm/s. Persistent force-tracking error pauses tangential
progress while bounded normal recovery continues. The robot releases contact
after reaching the endpoint. Table and foam retain the initial tool
orientation, whereas the phantom updates orientation using the estimated
local surface normal. Recovery pauses are part of the task and are included
in the elapsed scanning time.

Forces are gravity compensated and tared before each trial. The calibration
has a held-out $95$th-percentile force residual of $1.14$~N. Reported tracking
errors therefore characterize the compensated feedback signal rather than
independently verified absolute contact force, particularly for the low-force
foam condition.

\subsection{Surface-Scanning Results}
\label{app:real_surface_results}

We retain all five recorded trials for each surface. All $15$ scans reach
their path endpoint. Complete-session outcomes, including release and final
positioning, are $5/5$ for table, $4/5$ for foam, and $5/5$ for phantom.
The remaining foam session stops because robot feedback becomes stale after
scanning and release. Its completed scan remains included in the force
statistics.

The evaluation interval begins with tangential scanning and ends at scan
completion, including intermediate holds and contact recovery but excluding
initial approach, force settling, and final release. For compensated normal
force samples $F_{n,i}$, reference $F_n^\star$, and actual sampling intervals
$\Delta t_i$, we compute
\begin{equation}
\overline F_n=\frac{\sum_i\Delta t_iF_{n,i}}{\sum_i\Delta t_i}, \quad
\mathrm{MAE}=\frac{\sum_i\Delta t_i|F_{n,i}-F_n^\star|}{\sum_i\Delta t_i}, \quad
\mathrm{RMSE}=\sqrt{\frac{\sum_i\Delta t_i(F_{n,i}-F_n^\star)^2}
{\sum_i\Delta t_i}}.
\label{eq:real_force_metrics}
\end{equation}
Table~\visualref{tab:real_surface_repeats} reports the mean and sample standard
deviation of these per-trial quantities over five repetitions. The main-text
figure reports the same across-trial mean-force statistics.

\begin{table}[htbp]
\centering
\small
\setlength{\tabcolsep}{5pt}
\begin{tabular}{lcccc}
\toprule
Surface & $F_n^\star$ (N) & Mean $F_n$ (N) & MAE (N) & RMSE (N) \\
\midrule
Table & $3.0$ & $2.987\pm0.014$ & $0.391\pm0.054$ & $0.506\pm0.094$ \\
Foam & $0.8$ & $0.817\pm0.017$ & $0.176\pm0.006$ & $0.222\pm0.012$ \\
Phantom & $3.0$ & $2.940\pm0.064$ & $0.378\pm0.013$ & $0.454\pm0.015$ \\
\bottomrule
\end{tabular}
\caption{Surface-scanning results report the mean $\pm$ sample standard
deviation over five trials per material. Metrics use unsmoothed compensated
feedback and physical timestamps, including recovery intervals.
Complete-session outcomes are reported separately in the text.}
\label{tab:real_surface_repeats}
\end{table}

Mean force remains close to the reference on all three surfaces, but the
lower absolute error on foam does not imply easier regulation. Its mean
RMSE is approximately $27.8\%$ of the force reference, compared with
$16.9\%$ for table and $15.1\%$ for phantom, and its elapsed scanning time
is longer. The phantom additionally requires height and orientation
adaptation along a curved contact path. These results support coordinating
progress with measured contact response rather than executing a fixed-height
trajectory at constant speed. All reported metrics use unsmoothed
signals and physical elapsed time.

\subsection{Hole-Alignment and Insertion Results}
\label{app:real_insert_protocol}

The robot tracks a descending alignment reference from above the pipe rim
before entering the bore. Entry requires a robot-reported lateral offset below
$0.5$~mm and successful orientation checks. Insertion then follows the taught
axis with a $0.75$-mm offset limit. Axial and lateral force soft/hard
thresholds are $3/6$~N and $2/4$~N, respectively, with torque thresholds of
$0.5/1.0$~N\,m. These bounds govern the monitored execution, not a claim
of exact geometric clearance.

For pipe-axis location $(x_c,y_c)$ and robot-reported tool/flange reference
$(x(t),y(t))$, the lateral alignment measure is
\begin{math}
e(t)=\sqrt{(x(t)-x_c)^2+(y(t)-y_c)^2}.
\end{math}
In the recorded trial, pre-entry refinement reduces $e$ from $40.31$~mm to
$0.011$~mm at the rim. The robot then completes approximately $107$~mm of
insertion over $55.32$~s, with maximum in-pipe compensated axial and lateral
forces of $0.703$ and $1.329$~N. The offset is an internal robot-coordinate
measure, not an external measurement of shaft-to-wall clearance or absolute
mechanical accuracy.

Only one insertion trial is reported, and contact-force refinement is not
activated in that trial. It therefore demonstrates pre-entry geometric
alignment followed by monitored clearance insertion, rather than jam
recovery or sustained force-regulated insertion. Together with the repeated
surface trials, it provides deployment evidence for the model-based component
of \MGA{}. The contribution of the learned prior is evaluated separately
through the controlled simulation comparisons.

\section{Task-Specific Realization Maps}
\label{app:realization_maps}

The main text uses one residual $C^{\rm real}$ for all tasks. Here $t_0$
denotes the current control-step index and $h=0,\ldots,T-1$ the local prediction
index, corresponding to control step $t_0+h$. The implementation represents
$U$ by $N_u=H_{\rm node}+1$ control nodes of dimension $d_u$, so
$d_U=N_ud_u$. A fixed quadratic interpolation $\mathcal I$ produces
$U^{\rm dense}=\mathcal I U$ with $T=H_{\rm sample}+1$ controls.
Thus the dense sequence in the preliminaries is $U^{\rm dense}$, while the
method and theory operate on node coordinates with the rollout objective and
$C^{\rm real}$ composed with $\mathcal I$. Their derivatives include this
fixed interpolation. Let $u_h=(r_h,\upsilon_h^S,\nu_h)$ denote a dense control,
with $s_{\rm scale}\upsilon_h^S$ representing $S_h-S^{\rm ref}$ in symmetric
coordinates.

For the two arm tasks, the current realized-minus-command tracking error
$e_{t_0}$ and local response map $B_{t_0}$ define a frozen correction
\begin{math}
\delta_{t_0}=-B_{t_0}^\top(B_{t_0}B_{t_0}^\top+\lambda_BI)^{-1}e_{t_0},
\end{math}
where $\lambda_B>0$. The map is estimated by central differences of terminal
task coordinates under constant raw motion offsets over the first $H_B$
controls of the incumbent. It is $2\times2$ for surface coordinates and
$6\times6$ for insertion pose coordinates. The scanning error is the tangent
tracking error expressed in surface coordinates with the configured along-
and cross-path weights. Let $\bar\delta_{t_0}$ be the componentwise
bounded correction multiplied by the task reliability gate, which is one
when disabled. The horizon ramp is $b_h=\min\{(h+1)/H_B,1\}$.
Mapping $\bar\delta_{t_0}$ to task coordinates gives the target shifts below.
The current state, response map, gates, and target shifts are held fixed while
differentiating the chart. They change the clean chart rather than the
proposal bank.

For raw force coordinate $\nu$, the arm tasks use
$F_{\rm cmd}(\nu)=F_{\rm lo}
+\tfrac12(\operatorname{clip}_{[-1,1]}(\nu)+1)(F_{\rm hi}-F_{\rm lo})$,
where $F_{\rm lo}=F_{\min}-F_{\rm pad}$ and
$F_{\rm hi}=F_{\max}+F_{\rm pad}$ include the configured command padding.
Humanoid tasks use the same map with $F_{\rm pad}=0$, or the constant
$F_{\rm cmd}=F^\star$ when the force target is fixed. Here $F^\star$ is the
task's desired normal force. These maps are affine on each clipping branch.

\paragraph{Surface scanning.}
Let $q_{t_0}=(\xi_{t_0},\eta_{t_0})$ be the current commanded surface coordinate and let
$q_h(U)=\operatorname{clip}_{[0,1]^2}
(q_{t_0}+a_q\sum_{\ell=0}^{h}r_{\ell,1:2})$ be its cumulative horizon chart,
where $a_q$ converts raw motion into surface-coordinate increments. The shift
$\Delta_{t_0,h}^{q}$ is $b_h$ times $a_qH_B\bar\delta_{t_0}$ after the
configured correction gain and coordinate clipping. It defines
$\widetilde q_{t_0,h}^\star=q_{t_0,h}^\star+\Delta_{t_0,h}^{q}$.  For commanded-force
map $F_{\rm cmd}(\nu_h)$ and $s_F=F_{\max}-F_{\min}$, the residual is
\begin{equation}
C_{t_0}^{\rm scan}(U)
=\operatorname{stack}_h\left(
q_h(U)-\widetilde q_{t_0,h}^\star,
\frac{F_{\rm cmd}(\nu_h)-F^\star}{s_F}
\right).
\label{eq:app_surface_realization}
\end{equation}

\paragraph{Peg insertion.}
Let $p_{t_0}^{\rm cmd}$ be the current six-dimensional commanded pose and
$p_h(U)=p_{t_0}^{\rm cmd}+\sum_{\ell=0}^{h}D_pr_\ell$, where $D_p$ is the
diagonal map from raw translation and rotation channels to pose increments.
The response-aware target is
$\widetilde p_{t_0,h}^\star=p_{t_0,h}^\star+\Delta_{t_0,h}^{p}$, where
$p_{t_0,h}^\star$ is the approach--insertion reference and
$\Delta_{t_0,h}^{p}=b_hH_BD_p\bar\delta_{t_0}$. The chart uses diagonal
pose scaling $D_C$ with translation scales $(0.01,0.01,d_{\rm socket})$~m
and rotation scales $(0.1,0.1,0.1)$~rad, where $d_{\rm socket}$ is the socket
depth. With $s_F=\max\{F_{\max},1\}$, its contact activation is
$w_{t_0,h}=\sigma(\widetilde p_{t_0,h,z}^\star/s_z)$, where
$\sigma(x)=(1+e^{-x})^{-1}$ and $s_z=0.001$~m is the activation scale:
\begin{equation}
C_{t_0}^{\rm insert}(U)
=\operatorname{stack}_h\left(
D_C^{-1}\!\left[p_h(U)-\widetilde p_{t_0,h}^\star\right],
w_{t_0,h}\frac{F_{\rm cmd}(\nu_h)-F^\star}{s_F}
\right).
\label{eq:app_insert_realization}
\end{equation}

\paragraph{Humanoid pushing.}
The whole-body task uses direct contact-semantic coordinates rather than an
incremental end-effector response correction.  Define
$c_h^S:=\operatorname{svec}(S_h-S^{\rm ref})
=s_{\rm scale}\upsilon_h^S$ and
$\varphi_h:=(F_{\rm cmd}(\nu_h)-F^\star)/(F_{\max}-F_{\min})$. Its chart is
$C_{t_0}^{\rm push}=\operatorname{stack}_h C_{t_0,h}^{\rm push}$, with
$\Delta_{t_0}^{\rm real}=0$ and
\begin{equation}
C_{t_0,h}^{\rm push}(U)
=
\begin{cases}
[c_h^S,\varphi_h]^\top,
&\text{force regulation or fixed-stance pushing},\\
[c_h^S,\varphi_h,(r_h^{\rm contact})^\top]^\top,
&\text{unjamming},\\
[\varphi_h],
&\text{walking push},
\end{cases}
\label{eq:app_push_realization}
\end{equation}
Here $\operatorname{svec}$ vectorizes a symmetric matrix and
$r_h^{\rm contact}$ denotes the face-selection logits. The remaining
contact coordinates are optimized through rollout evaluation rather than
fixed by this chart. Measured contact reliability is recorded separately,
and the direct clean chart is unchanged.  In the fixed-force suites,
$\varphi_h$ is identically zero.  It may be omitted in the local constant-rank
interpretation, although the implementation retains the row and handles its
redundancy through regularization.

\paragraph{Soft candidate evaluation.}
\label{app:soft_candidate_evaluation}
For the soft evaluation cost in Section~\ref{subsec:prelim_joint_inference}, let $v^h\geq0$ collect task-selected
equality magnitudes and positive inequality violations along the rollout.
Following MD-COAS~\citep{he2026motion}, we use
\begin{math}
\mathcal V(U;s^0,c)
=\sum_{h=0}^{T-1}\left(\lambda^\top v^h
+\tfrac{\rho}{2}\lVert v^h\rVert_2^2\right),
\end{math}
with $\lambda\succeq0$ and $\rho\geq0$.  These evaluation residuals are
distinct from the realization charts above.

\section{Additional Results}
\label{app:additional_results}

We examine realized path fidelity, productive insertion, and continued
whole-body interaction. All three environments use seeds $0$--$9$, with
the evaluation rules in Appendix~\appref{app:env_setup}.

\subsection{Surface Scanning}
\label{app:additional_scanning}

\begin{figure}[!htbp]
\centering
\includegraphics[width=\linewidth]{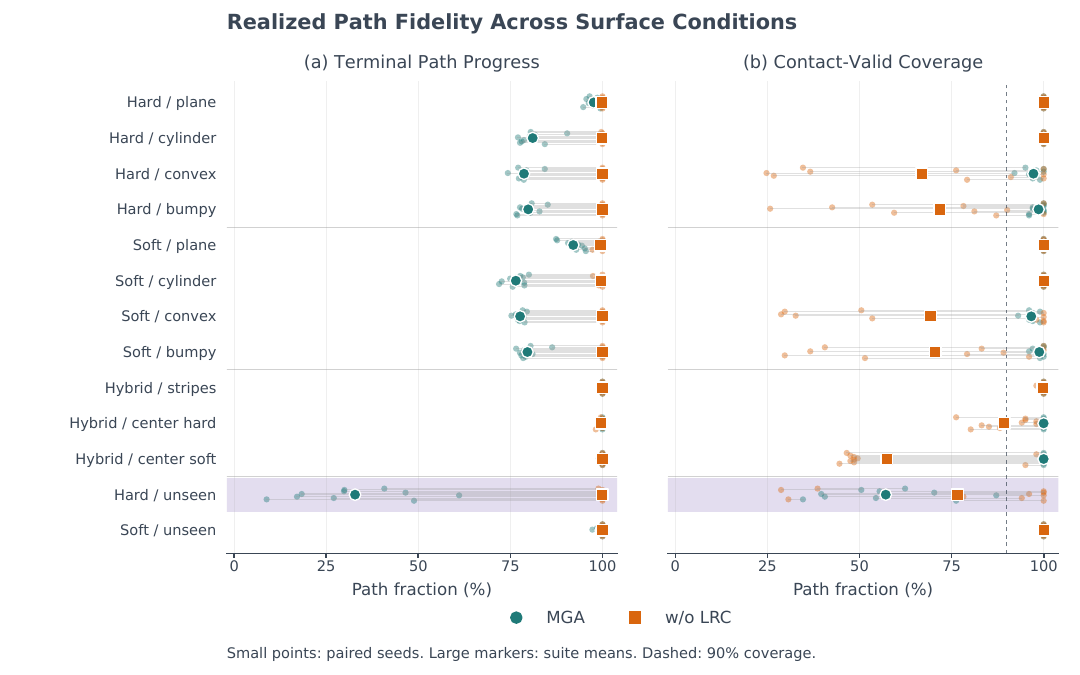}
\caption{The plots compare terminal progress and contact-valid coverage across all
scanning conditions with matched settings and seeds. Large markers show suite
means, and small markers retain individual executions. Held-out surfaces are not
included again in the Hard and Soft groups.}
\label{fig:app_scan_coverage}
\end{figure}

\paragraph{Realized coverage beyond terminal progress.}
Removing LRC yields $99.88\%$ mean terminal progress, but reduces
contact-valid coverage from $96.03\%$ to $84.75\%$ across the $13$ conditions
(Fig.~\visualref{fig:app_scan_coverage}). Soft-center coverage falls from
$100\%$ to $57.43\%$. Terminal progress projects the final position onto
the reference, whereas coverage counts reference waypoints visited with
valid contact within $5$~mm. Their disagreement shows why advancing the
endpoint is insufficient: correction must preserve the executed path.

\paragraph{Sensitivity to the success definition.}
Disallowing the operational criterion's single force-violating sample
leaves the coverage benefit intact (Table~\visualref{tab:app_scan_success_check}).
Of the $43$ no-LRC failures, $42$ fail coverage alone and one fails settled
contact: path fidelity, rather than the force allowance, dominates.

\begin{table}[!htbp]
\centering
\small
\setlength{\tabcolsep}{9pt}
\begin{tabular}{lccc}
\toprule
Method & Operational success & Zero-violation success & Coverage (\%)\\
\midrule
\MGA{} & $120/130$ & $119/130$ & $96.03\pm1.38$\\
w/o RL prior & $120/130$ & $117/130$ & $97.10\pm1.26$\\
w/o LRC & $87/130$ & $86/130$ & $84.75\pm6.19$\\
\bottomrule
\end{tabular}
\caption{Scanning success is evaluated under two force-violation allowances.
Both criteria require coverage $\geq90\%$ and settled contact loss
$\leq10\%$. Counts cover all $13\times10$ executions. Coverage is reported as the
mean and population standard deviation of the ten equal-suite seed means.}
\label{tab:app_scan_success_check}
\end{table}

\begin{figure}[!htbp]
\centering
\includegraphics[width=\linewidth]{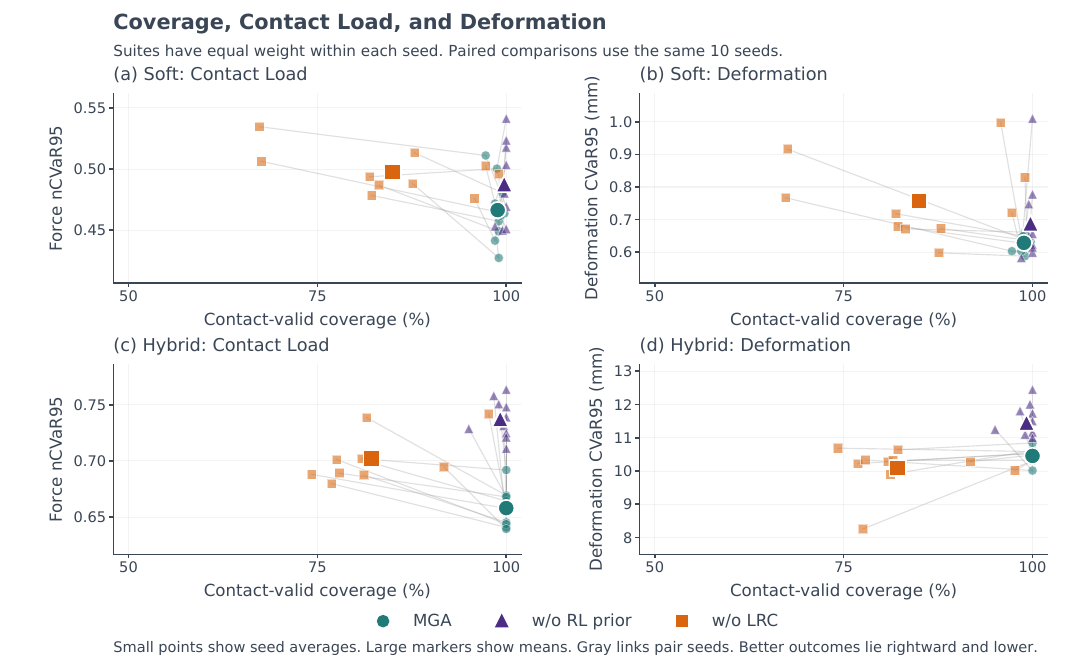}
\caption{Matched seed-level observations compare coverage, force tails, and
deformation for MGA and variants without the learned prior or LRC under
compliant and hybrid contact. Force utilization is normalized by the $60$-N
limit, and deformation is reported in physical units.}
\label{fig:app_scan_tradeoff}
\end{figure}

\paragraph{Progress, contact load, and deformation.}
On Soft Convex, MGA reduces force-tail utilization
from $0.432$ to $0.388$ and deformation CVaR$_{95}$ from $0.548$ to
$0.493$~mm relative to w/o LRC while recovering coverage
(Fig.~\visualref{fig:app_scan_tradeoff}). Adding the prior also lowers
Hybrid force tails from $0.738$ to $0.658$. These benefits are conditional:
on unseen rigid contact, removing LRC improves coverage from $57.13\%$ to $76.63\%$.

\begin{figure}[!htbp]
\centering
\includegraphics[width=\linewidth]{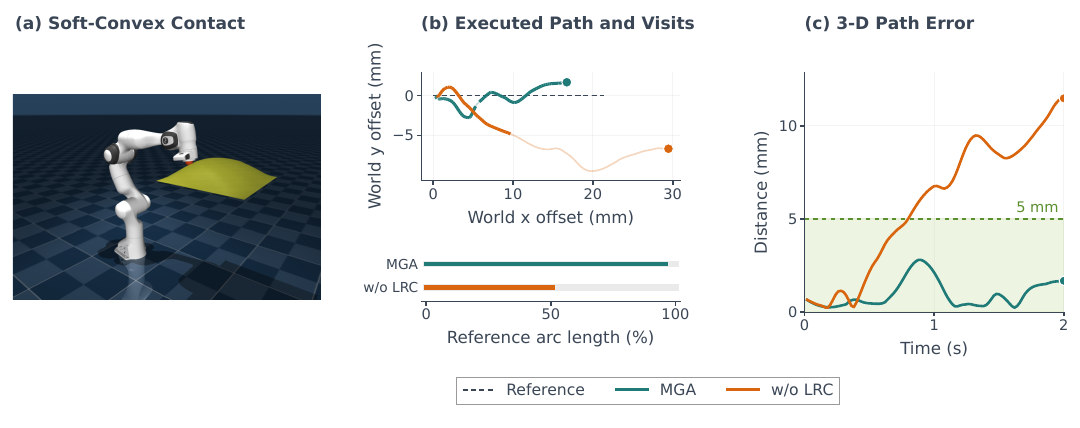}
\caption{The Soft Convex panels show native MGA context and matched executions.
Visit strips count reference waypoints within $5$~mm of contact-valid
executed samples in 3-D. The error trace measures executed-point distance
to the reference polyline, also in 3-D. The boundary marks the same tolerance.}
\label{fig:app_scan_execution_geometry}
\end{figure}

\paragraph{Spatial evidence of realized path fidelity.}
Figure~\visualref{fig:app_scan_execution_geometry} exposes the execution-level
effect on a compliant curved surface. MGA visits $96.04\%$ of the path with
valid contact, versus $50.50\%$ without LRC, although terminal progress
is $78.33\%$ and $100\%$, respectively. The no-LRC execution advances
while leaving the tracking tolerance, rather than faithfully covering the path.

\begin{figure}[!htbp]
\centering
\includegraphics[width=\linewidth]{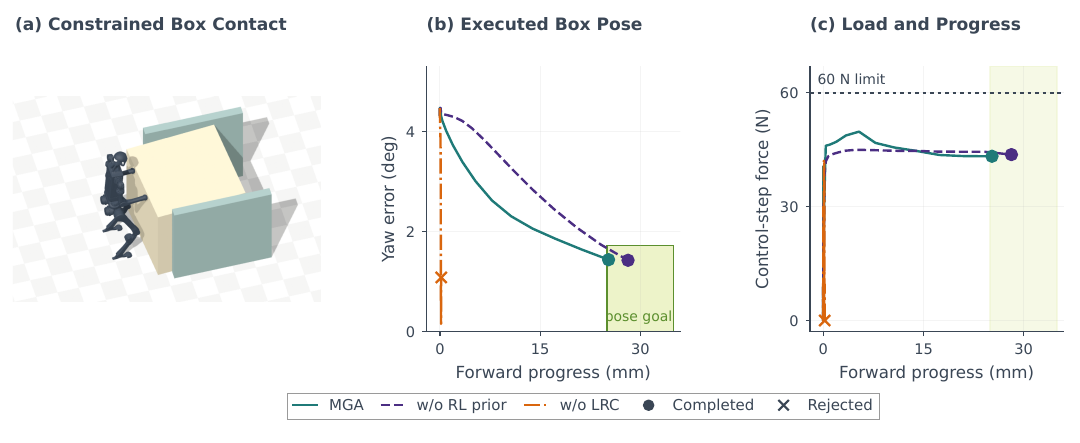}
\caption{The panels show simulated Unjamming behavior for MGA, w/o RL prior,
and w/o LRC. Native imagery provides task context. The plotted motion and
force are from the matched executions, ending at completion or
rejection without absorbing padding. Force is synchronized with the
$20$-ms control-step pose records. Physical impact and safety analyses
elsewhere retain the finer $4$-ms samples. The position--yaw target is a
task-completion condition, not a complete safety certificate.}
\label{fig:app_humanoid_execution_geometry}
\end{figure}

\begingroup
\setlength{\intextsep}{5pt}
\setlength{\textfloatsep}{8pt}
\subsection{Peg Insertion}
\label{app:additional_peg_insert}
\label{app:peg_insert_results}

\paragraph{Completion and contact safety.}
Reaching the target and completing safely are separate outcomes
(Fig.~\visualref{fig:app_peg_outcomes}). The contact utilization is
\begin{equation}
\rho_t=\max\left\{
\frac{\lVert F_{xy,t}\rVert}{20\,\mathrm{N}},
\frac{|F_{z,t}|}{30\,\mathrm{N}},
\frac{\lVert\tau_{xy,t}\rVert}{1.5\,\mathrm{N\,m}},
\frac{|\tau_{z,t}|}{1.0\,\mathrm{N\,m}}
\right\}.
\label{eq:app_peg_utilization}
\end{equation}
We evaluate the recorded simulated loads even under sensing bias.
For each $64$-step execution, $\operatorname{nCVaR}_{95}$ averages the four
largest values of $\rho_t$. Safe completion requires success with no wrench
violation or jam in this complete window, including after first completion.
\MGA{} completes $28/30$ trials safely, compared with $23$ for PegasusFlow,
$17$ each for MPPI and DIAL, $5$ for ATACOM, and $0$ for ISSA.

\begin{figure}[!htbp]
\centering
\includegraphics[width=\linewidth]{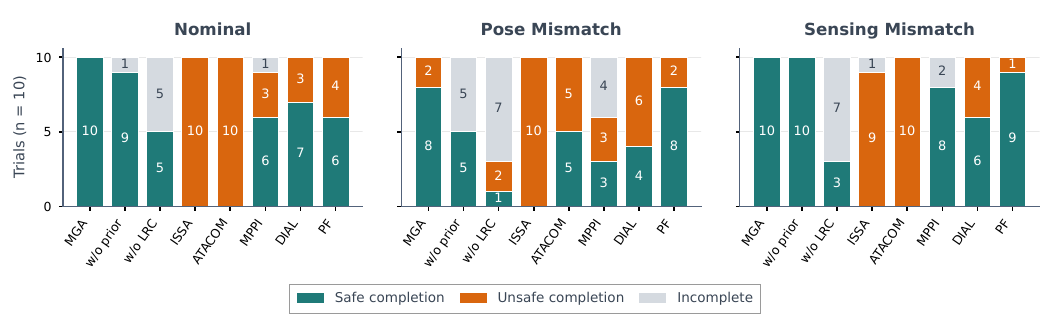}
\caption{Insertion outcomes are reported across ten seeds per method and setting.
All eight methods use the same frozen evaluation conditions. Safe completion,
completion with a wrench violation or jam, and incomplete insertion are
reported separately. No trial is discarded.}
\label{fig:app_peg_outcomes}
\end{figure}

\paragraph{When learned behavior supports productive insertion.}
Under Pose-OOD, the prior raises raw completion from $5/10$ to $10/10$
and safe completion from $5/10$ to $8/10$, while mean tail utilization
rises from $0.769$ to $0.893$. Under ID it raises safe completion from
$9/10$ to $10/10$. Under Sensing-OOD both variants complete $10/10$ safely.
Figure~\visualref{fig:app_peg_response} retains incomplete trials to distinguish
low-load stalling from successful execution.

\begin{figure}[H]
\centering
\includegraphics[width=\linewidth]{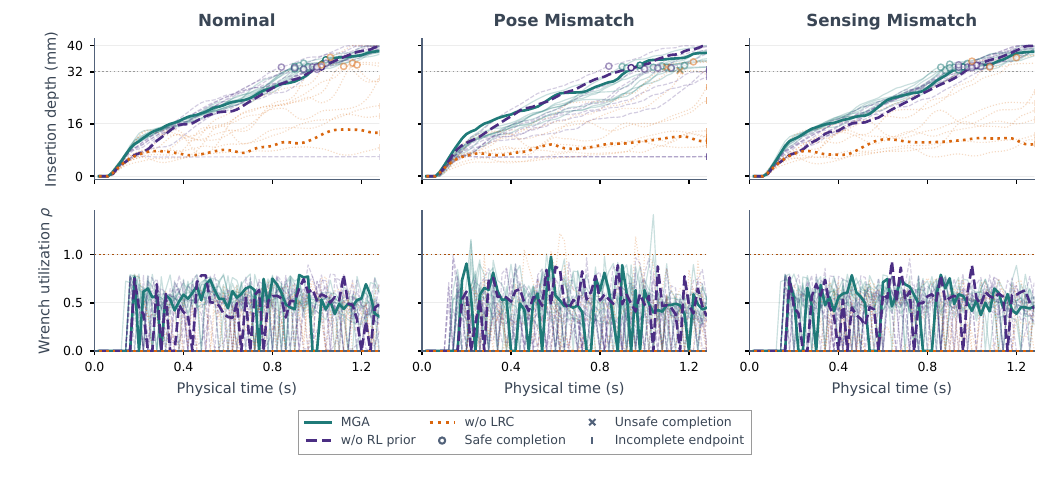}
\caption{Depth and wrench utilization for ten seeds per variant and setting,
with seed $0$ emphasized. All recorded steps are retained. The depth and
$\rho=1$ thresholds do not subsume pose tolerances or the three-step hold.}
\label{fig:app_peg_response}
\end{figure}

\begin{table}[!htbp]
\centering
\small
\setlength{\tabcolsep}{4pt}
\renewcommand{\arraystretch}{1.08}
\begin{tabular}{llcccc}
\toprule
Setting & Method & Completed & $F_{xy}^{\max}$ (N) & $F_z^{\max}$ (N)
& F/T violation (\%)\\
\midrule
\multirow{3}{*}{ID}
& \MGA{} & $10/10$ & $15.85\pm0.13$ & $14.51\pm1.18$ & $0.00\pm0.00$\\
& w/o RL prior & $9/10$ & $15.14\pm1.93$ & $15.04\pm1.12$ & $0.00\pm0.00$\\
& w/o LRC & $5/10$ & $10.46\pm6.14$ & $7.98\pm5.05$ & $0.00\pm0.00$\\
\midrule
\multirow{3}{*}{Pose-OOD}
& \MGA{} & $10/10$ & $20.05\pm3.09$ & $18.00\pm3.48$ & $0.62\pm1.25$\\
& w/o RL prior & $5/10$ & $17.17\pm3.02$ & $16.73\pm2.46$ & $0.00\pm0.00$\\
& w/o LRC & $3/10$ & $12.46\pm9.05$ & $10.01\pm8.23$ & $0.78\pm1.88$\\
\midrule
\multirow{3}{*}{Sensing-OOD}
& \MGA{} & $10/10$ & $16.04\pm0.58$ & $14.40\pm1.22$ & $0.00\pm0.00$\\
& w/o RL prior & $10/10$ & $16.52\pm1.25$ & $15.77\pm2.02$ & $0.00\pm0.00$\\
& w/o LRC & $3/10$ & $9.57\pm6.42$ & $7.79\pm5.70$ & $0.00\pm0.00$\\
\bottomrule
\end{tabular}
\caption{The table reports contact outcomes for \MGA{} and both controlled ablations.
Completed counts denote raw task completion. Loads and violation rates
are mean $\pm$ population standard deviation over all ten trials, including
failures. None of these three variants records a jam.}
\label{tab:app_peg_contact}
\end{table}

\paragraph{Geometric mismatch and evidence scope.}
Without LRC, only $11/30$ insertions complete and $9/30$ complete safely.
Its lower mean tail utilization ($0.453$ versus $0.814$) accompanies a
smaller terminal depth ($23.2$ versus $37.9$~mm), so low load alone cannot
establish effective contact control.
The two unsafe \MGA{} trials occur under Pose-OOD: lateral-force peaks
reach $28.31$ and $22.81$~N against the $20$-N limit. PegasusFlow matches
its $8/10$ safe completions in this setting and attains lower mean tail
utilization ($0.630$ versus $0.893$). Thus, the overall gain does not imply
uniform dominance or absolute safety under model mismatch.
All settings share one additive-prior configuration and apply execution
perturbations to every method. ATACOM inserts in all $30$ trials but violates
wrench limits in $25$. ISSA inserts in $29$, with $25$ jams and $11$ wrench
violations (overlapping counts).

\paragraph{Alignment and contact load along the insertion.}
Figure~\visualref{fig:app_peg_geometry_contact} relates the replayed insertion
to lateral-error--depth paths, retaining stalled and unsafe executions.
The positional target does not certify contact safety. The full-window
criterion also checks orientation, hold time, and load history.

\begin{figure}[H]
\centering
\includegraphics[width=\linewidth]{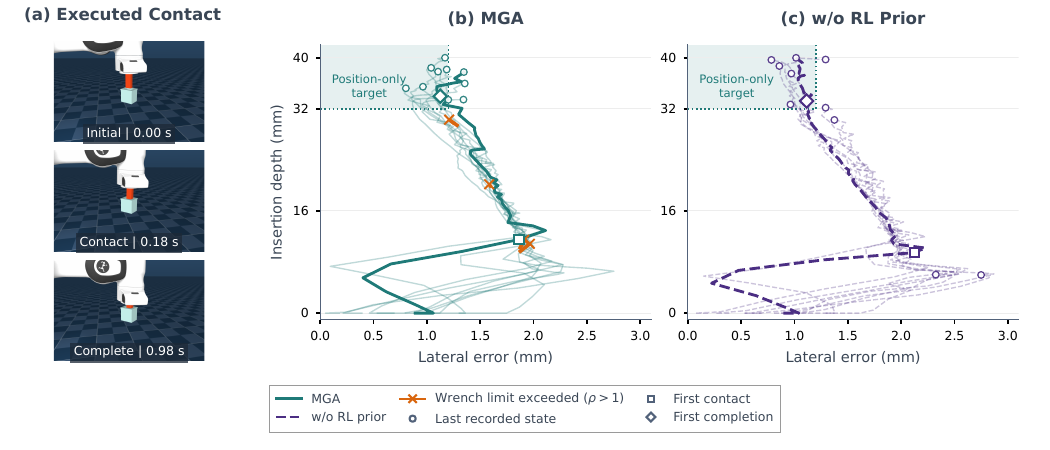}
\caption{Pose-OOD replay and all ten recorded executions for \MGA{} and w/o RL
prior, with seed $0$ emphasized. Circles mark endpoints and diamonds first
completion. Load exceedances remain visible. The shaded positional target
uses lateral error $\leq1.2$~mm and depth $\geq32$~mm, separately from the
orientation, hold-time, and full-window safety conditions.}
\label{fig:app_peg_geometry_contact}
\end{figure}

\endgroup
\begingroup
\setlength{\intextsep}{5pt}
\subsection{Humanoid Push}
\label{app:additional_humanoid}

\paragraph{Force transients and tracking.}
MGA's mean impact peak is $23.46\pm4.02$~N, compared with
$48.63\pm3.93$~N for MPPI, $49.78\pm8.74$~N for DIAL, and
$52.73\pm7.55$~N for PegasusFlow. MGA reduces impact but does not minimize every tracking
metric (Fig.~\visualref{fig:app_humanoid_force}). The $4$-ms physics samples resolve peaks potentially missed by
control-step observations.

\begin{figure}[!htbp]
\centering
\includegraphics[width=0.96\linewidth]{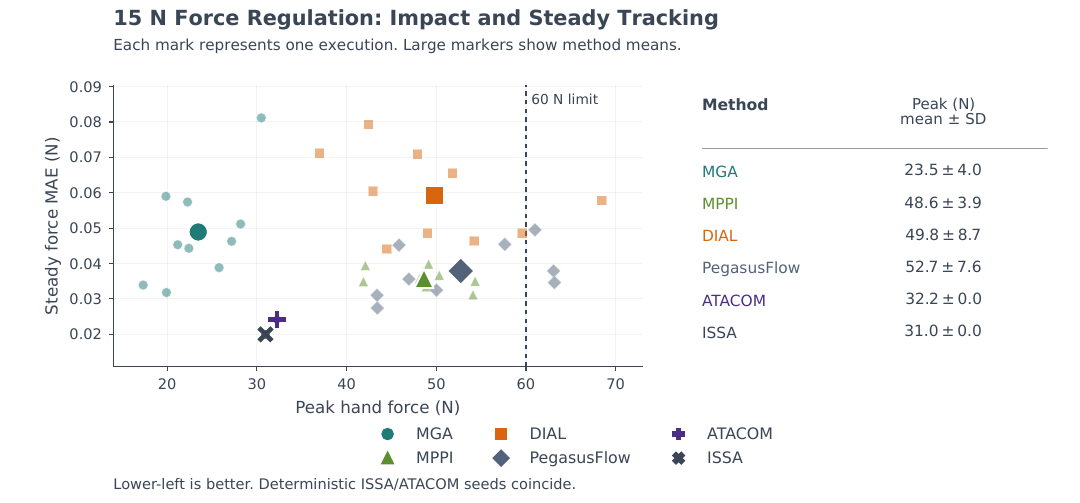}
\caption{The plot compares impact and steady-state error for $15$~N Force Regulation.
Small points denote seeds, and large markers denote method means.
All methods share physical sampling.}
\label{fig:app_humanoid_force}
\end{figure}

\paragraph{Constructing continued interaction.}
In Unjamming, MGA safely completes $10/10$ trials, versus $8/10$
without the prior and $3/10$ without LRC
(Fig.~\visualref{fig:app_humanoid_continuation}). All seven no-LRC failures
stop because no candidate passes revalidation. Correction sustains
interaction under the same execution checks.

\begin{figure}[!htbp]
\centering
\includegraphics[width=0.96\linewidth]{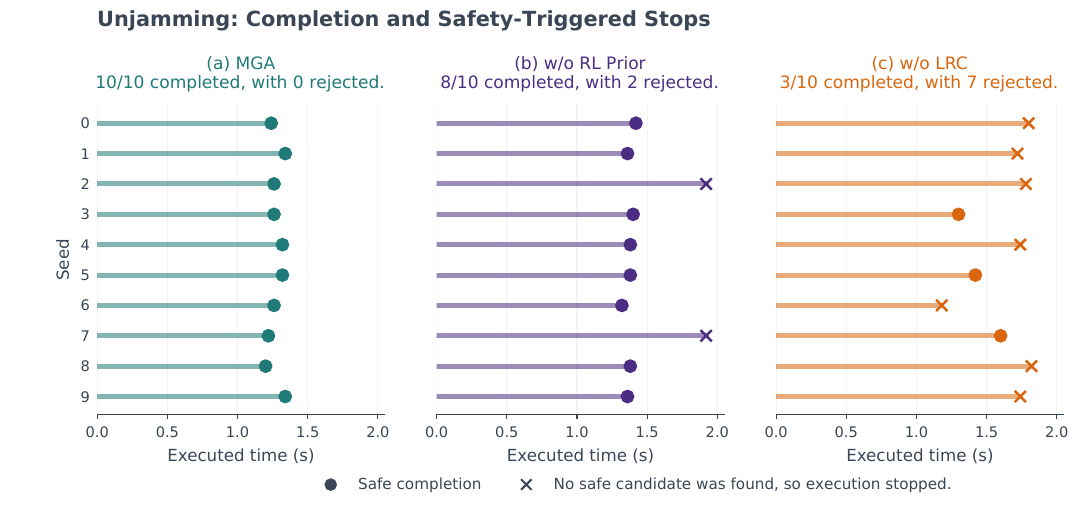}
\caption{Unjamming traces end at completion or safety-triggered rejection, without
absorbing padding. Rejected actions are not executed.}
\label{fig:app_humanoid_continuation}
\end{figure}

\paragraph{Load regulation beyond the fixed-force task.}
All three Fixed-stance Push variants complete $10/10$ trials.
Force tails are $0.503$, $0.492$, and $0.538$ for MGA, w/o prior,
and w/o LRC. Prior benefits are task-dependent.
Unjamming tail summaries exclude rejected prefixes, which count as failures.
Conditional tails from eight no-prior and three no-LRC runs are not ranked
against MGA's ten-run estimate.

\endgroup
\paragraph{Motion before completion or rejection.}
Figure~\visualref{fig:app_humanoid_execution_geometry} connects the stopping
events to the box motion. Translation and yaw must be corrected together.
A bounded interaction force with unresolved object error does not constitute
task completion. The paired paths make this distinction visible while retaining
the actual rejection endpoint of w/o LRC. Together with the all-seed
continuation counts, this supports constructing useful accepted sequences
rather than interpreting safe rejection alone as successful control.

\section{Proposal and Refinement Details}
\label{app:proposal_implementation}

This section specifies the proposal protocols in
Section~\ref{subsec:method_rl_prior}. We fix the current state $s^0$ and context
$c$. The index $h$ denotes local physical rollout time, and $m$ indexes samples.
Solver and dense coordinates follow Appendix~\appref{app:realization_maps}.

\paragraph{Proposal coordinates and reuse.}
Learned horizons are mapped to the solver coordinates. Gaussian proposals use
the DIAL-style node-dependent annealing scale around $\check U_k$, with the
first node fixed. Control clipping and any enabled trust map are applied before
evaluation, so this implemented proposal is not the unmodified isotropic kernel
in Eq.~\eqref{eq:clean_proposal_U}. In replacement integration, learned
horizons occupy $M_\psi$ slots in the common $M$-candidate bank. They are drawn
once per solve and their coordinate and trust maps are reapplied at each
reverse step. Gaussian proposals are refreshed at every step.

In additive integration, $M_\psi=0$ in this refinement bank. The Gaussian
search starts from the shifted incumbent with the same random stream and
realization context as the no-prior variant. After its refinement and
execution revalidation, one deterministic and eight stochastic policy
horizons are evaluated separately. The best admissible expert competes with
the selected Gaussian-path output under the acceptance rules below. These
expert horizons do not enter the common softmax or undergo that solve's
geometric shaping and LRC. Their evaluations are additional to $M$.

\paragraph{Optional constraint-aware policies.}
In replacement integration, the learned budget $M_\psi$ can be split across an RL policy and a stateful
constraint-aware expert such as ATACOM. For the latter, let $b_m^h$ be its free
action and $\zeta_m^h$ its hypothetical slack state:
\begin{equation}
b_m^h\sim\pi_{\rm at}(\cdot\mid o_m^h),\quad
(u_m^h,\zeta_m^{h+1})
=\mathcal T_{\rm at}(s_m^h,b_m^h,\zeta_m^h), \quad
U_m^{\rm at}\sim Q_{\rm at}(\cdot\mid s^0,\zeta^0,c).
\label{eq:atacom_horizon_prior}
\end{equation}
Each rollout evolves its own slack copy initialized at $\zeta^0$. The persistent
slack is advanced only after real execution. Expert slots replace other
stochastic slots rather than increasing $M$.

\paragraph{Initialization and localization.}
Let $U^\psi$ denote the deterministic policy rollout. The reverse process starts
from $U_K=\sqrt{\bar\alpha_K}U^{\rm init}$, where
\begin{equation}
U^{\rm init}
=\lambda_{\rm shift}U^{\rm shift}
+(1-\lambda_{\rm shift})U^\psi,
\quad \lambda_{\rm shift}\in[0,1].
\label{eq:rl_warm_start}
\end{equation}
This initialization applies to replacement integration. Additive integration
uses $U^{\rm init}=U^{\rm shift}$ and does not localize Gaussian proposals
around the policy. At the first replan, $U^{\rm shift}$ is zero-initialized.
When enabled in replacement integration, proposal localization maps sequences into
\begin{equation}
\left\{U:\min_{U^{\rm ref}\in\mathcal C}
\frac{\lVert U-U^{\rm ref}\rVert_2}{\sqrt{d_U}}
\leq\epsilon_{\rm trust}\right\},
\label{eq:union_trust_region}
\end{equation}
where $U^{\rm ref}$ ranges over the reference set $\mathcal C=\{U^\psi\}$,
optionally enlarged by a deterministic expert rollout, and $d_U$ is the
number of optimized scalar coordinates.
The trust map is also applied after transport, before residual correction.
It does not guarantee that anchors or corrected sequences remain in this set.

\paragraph{Weight normalization.}
Equation~\eqref{eq:method_clean_mean} uses $\beta_k=\tau\sigma_{J,k}$, where
$\tau>0$ is the sampling temperature and $\sigma_{J,k}$ is the cost standard
deviation over all $N_k$ evaluated candidates, replaced by one below the numerical
threshold $10^{-6}$. Subtracting the clean iterate's cost from every softmax logit, as in
the implementation, leaves the normalized weights unchanged.

\paragraph{Encoded geometry and modulation.}
\label{app:geometry_implementation}
For $\Phi(U):=\lVert C^{\rm real}(U)\rVert_2^2/(2n)$, where $n$ is the
residual dimension, let $a_j\in\mathbb R^{d_u}$ be node $j$'s block of
$\nabla_U\Phi(\check U_k)$. Select the $K_a$ blocks with largest Euclidean
norm, where $K_a=\min\{K_{\rm configured},N_u\}$, and use
$a_j^\top/(\lVert a_j\rVert_2+\epsilon)$ as the rows of
$A_k^{\rm node}\in\mathbb R^{K_a\times d_u}$. Here $\epsilon>0$ is the
numerical stabilizer. A zero block therefore remains a zero row.
The sequence-space operator is $A_k=I_{N_u}\otimes A_k^{\rm node}$,
with $G_k=I$ and $\lambda_k=1+\epsilon$.
At $C^{\rm real}=0$, this energy gradient vanishes even when
$\mathsf J\neq0$, so Lemma~2 alone does not imply contraction of every
true chart-normal response.

The task-dependent implementation uses
$d_k^{\rm impl}=d_k+\Gamma(\chi_k\mathsf M_kd_k-d_k)$ and transports the
clean estimate $\check U_k+d_k^{\rm impl}$. The diagonal matrix $\Gamma$
repeats the task's action-coordinate reliability gates across the $N_u$
nodes. Each gate lies in $[0,1]$, and an ungated clean block uses one.
The scalar $\chi_k$ is the configured geometric gain multiplied by the
adaptive feasibility factor, which is one when adaptive modulation is off.
The normalized core analyzed in the main text sets $\Gamma=I$ and
$\chi_k=1$. For $\Gamma=\gamma I$, $0\leq\gamma\leq1$, and $\chi_k=1$,
the filter's eigenvalues become
$1-\gamma+\gamma\lambda_k/(\lambda_k+\sigma_i^2)$, where $\sigma_i$ are the
singular values of $A_k$, so nonexpansiveness and kernel preservation remain
valid. Arbitrary coordinate-dependent $\Gamma$ or gain $\chi_k$ need not
inherit these properties.

\paragraph{Residual correction.}
Equation~\eqref{eq:local_residual_correction} acts on the clean sequence after
transport and any enabled trust mapping. If repeated, it recomputes the full
chart Jacobian and residual at each corrected sequence. Task-dependent
reliability can also modulate this correction. Theorem~1 concerns the
unmodulated Jacobian correction with $0<\eta\leq1$. Its correction-only
bound still holds for a trust-mapped input on the same affine branch, but
the composed identity from $\check U_k$ requires the stated normalized
transport. Subsequent task projections and execution acceptance are separate
operations and are not covered by these local contraction claims.

\paragraph{Execution revalidation and fallback.}
Each enabled acceptance check reevaluates the refined sequence and incumbent
from the current measured state. The task predicate checks predicted normal-
force violations for scanning, force/torque violations and jamming for
insertion, and force, contact-validity, and balance violations for humanoid
pushing. Their definitions follow Appendix~\appref{app:env_setup}.
If both performance candidates pass, refinement must improve the rollout
score beyond the configured tolerance without worsening the task's protected
risk components beyond their tolerances. If only refinement passes, it can
replace an invalid incumbent without a score improvement. Otherwise a valid
incumbent is retained. When neither passes, or a task recovery override is
active, the controller considers the task's emergency candidates.

Additive experts first pass the same task predicate and any applicable
reliability checks. The best-scoring admissible expert is tested against the
already selected Gaussian-path output. Its adoption requires acceptance and
strict score improvement on ordinary replans. Task-owned recovery transitions
use their dedicated rules and cannot be overwritten by a policy score win.
Rejection retains the Gaussian-path decision. For the humanoid execution
context, failure of all revalidated continuations stops the run before
executing the rejected action. These checks explain the reported execution
rejections but do not certify physical safety under model mismatch.

\begingroup
\setlength{\intextsep}{5pt}
\setlength{\textfloatsep}{8pt}
\section{Hyperparameters}
\label{app:hyperparameters}

The settings below follow saved configurations and frozen-component metadata.
Environment geometry, physical limits, and success definitions appear in
Appendix~\appref{app:env_setup}.

\subsection{Learned Proposal Settings}
\label{app:prior_hparams}

All proposal policies remain frozen during evaluation.
Table~\visualref{tab:app_prior_hparams} separates training budgets from selected
checkpoints. Scanning also uses a separately trained ATACOM proposal expert.
The comparisons test deployment of learned behavior, not further training.

\begin{table}[!htbp]
\centering
\small
\setlength{\tabcolsep}{5pt}
\renewcommand{\arraystretch}{1.12}
\begin{tabular}{p{0.31\linewidth}ccc}
\toprule
Setting & Scanning & Peg Insertion & Humanoid Push\\
\midrule
Policy optimizer & PPO & PPO & PPO\\
Hidden layers & $4\times32$ & $2\times16$ & $2\times16$\\
Recorded training budget & $2$ M & $200$ k & $200$ k\\
Selected PPO checkpoint & $1$ M & $200$ k & Final\\
ATACOM proposal expert & Yes & No & No\\
Selected ATACOM checkpoint & $500$ k & -- & --\\
Policy updates in evaluation & None & None & None\\
\bottomrule
\end{tabular}
\caption{The table lists frozen proposal settings, with budgets measured in
training environment steps. The ATACOM rows describe a proposal expert. The separate baseline
is described in Appendix~\appref{app:baseline_details}.}
\label{tab:app_prior_hparams}
\end{table}

The scanning policy trains on rigid cylinder and convex surfaces and compliant
cylinder and bumpy surfaces. The insertion policy uses nominal, tight-clearance,
and randomized conditions.
H1 shares one policy across its three tasks. Checkpoints follow the saved
development criteria: realized progress for scanning, the final fixed-budget
checkpoint after an independent health check for insertion, and the final
fixed-budget policy for H1. The separate insertion ATACOM baseline also
uses its final $200$-k-step tangent-policy checkpoint.

\subsection{Model-Based Annealing Settings}
\label{app:annealing_hparams}

Table~\visualref{tab:app_annealing_hparams} separates evaluation length from
planning horizon. Table~\visualref{tab:app_proposal_budgets} records the
proposal sources needed to interpret rollout work.

\begin{table}[!htbp]
\centering
\small
\setlength{\tabcolsep}{6pt}
\renewcommand{\arraystretch}{1.10}
\begin{tabular}{p{0.38\linewidth}ccc}
\toprule
Setting & Scanning & Peg Insertion & Humanoid Push\\
\midrule
Evaluation intervals & $100$ & $64$ & $100$\\
Control interval (s) & $0.02$ & $0.02$ & $0.02$\\
Dense-grid intervals ($H_{\mathrm{sample}}$) & $16$ & $8$ & $16$\\
Node intervals ($H_{\mathrm{node}}$) & $4$ & $4$ & $4$\\
Refinement-bank budget $M$ & $64$ & $64$ & $64$\\
Initial refinement iterations & $10$ & $4$ & $10$\\
Subsequent refinement iterations & $2$ & $2$ & $2$\\
Cost-scaled temperature $\tau$ & $0.06$ & $0.06$ & $0.06$\\
\bottomrule
\end{tabular}
\caption{The table lists the recorded MGA inference settings. Peg insertion retains
the full $64$-step window after completion. Other tasks use their documented
termination rules. Sampling temperature enters $\beta_k=\tau\sigma_{J,k}$
and is not a physical contact parameter.}
\label{tab:app_annealing_hparams}
\end{table}

\begin{table}[!htbp]
\centering
\small
\setlength{\tabcolsep}{4pt}
\renewcommand{\arraystretch}{1.10}
\begin{tabular}{lcccc}
\toprule
Recorded setting & Gaussian & PPO & ATACOM & Integration\\
\midrule
Scanning & $48$ & $8$ & $8$ & Replacement bank\\
Peg Insertion: all settings & $64$ & $8$ & $0$ & Additive expert\\
Humanoid Push & $64$ & $8$ & $0$ & Additive expert\\
\bottomrule
\end{tabular}
\caption{The table lists the recorded stochastic proposal counts. Deterministic anchors
and execution-revalidation queries are additional.}
\label{tab:app_proposal_budgets}
\end{table}

Replacement uses learned slots in the $64$-sample bank
(Section~\ref{subsec:method_rl_prior}), reusing those proposals across reverse
steps while refreshing Gaussian samples. Additive integration retains
$64$ Gaussian samples and separately evaluates one deterministic plus eight
stochastic policy horizons after Gaussian refinement and revalidation.
These nine horizons are outside the softmax bank and its shaping/LRC updates.
Peg ID and OOD use the same additive configuration.
The prior comparison therefore retains additional policy rollouts and does
not establish equal-total-rollout efficiency. Action-filtering baselines
do not inherit a trajectory sampler.

\subsection{Realization-Aware Refinement Settings}
\label{app:refinement_hparams}

Task-specific charts (Appendix~\appref{app:realization_maps}) share the
operators in Table~\visualref{tab:app_refinement_hparams}. Feasibility
penalties evaluate candidates, and LRC corrects the transported clean sequence.

\begin{table}[!htbp]
\centering
\small
\setlength{\tabcolsep}{6pt}
\renewcommand{\arraystretch}{1.10}
\begin{tabular}{p{0.38\linewidth}ccc}
\toprule
Setting & Scanning & Peg Insertion & Humanoid Push\\
\midrule
Linear feasibility coefficient & $2$ & $2$ & $2$\\
Quadratic feasibility coefficient & $200$ & $80$ & $200$\\
Normalized geometric metric & $I$ & $I$ & $I$\\
LRC iterations per correction & $1$ & $1$ & $1$\\
LRC gain $\eta$ & $1$ & $1$ & $1$\\
LRC regularization $\varepsilon$ & $10^{-6}$ & $10^{-6}$ & $10^{-6}$\\
\bottomrule
\end{tabular}
\caption{The table lists the evaluation and normalized correction settings. Task modulation,
trust maps, and execution checks are detailed in
Appendix~\appref{app:proposal_implementation}. The operator alone does not certify safety.}
\label{tab:app_refinement_hparams}
\end{table}

The \textbf{w/o RL prior} variant retains the task's refinement and execution mechanisms.
The \textbf{w/o LRC} variant retains proposals, geometric shaping, and acceptance checks
and does not remove physical withdrawal. PegInsert uses core-only refinement
with learned reliability disabled in all three variants. In scanning and H1,
learned reliability defers to task-owned checks, so these ablations do not
establish an independent learned-confidence gain or a continuous parameter sweep.

\endgroup
\clearpage
\begingroup
\setlength{\intextsep}{5pt}
\makeatletter
\setlength{\@fptop}{0pt}
\makeatother
\section{Visualizations}
\label{app:visualizations}

Each row shows five states from one MGA execution, with a fixed
camera and native robot assets. We retain all $13$ scanning suites, three
insertion settings, and three humanoid tasks---including incomplete cases.
Quantitative analyses use all evaluation seeds.
Each frame reports simulation time $t$ and executed contact force $F$:
surface-normal force for scanning, contact-force magnitude for insertion,
and summed hand--box normal forces for humanoid tasks.
A dash indicates an unrecorded initial force, not zero force.

\subsection{Surface Scanning}
\label{app:vis_scanning}

Stiffness insets show nominal task-chart fields or uniform contact
parameters, not measured deformation. Rigid contact is categorical.
White dots locate the displayed command states, not end-effector world
positions, and the dashed line marks $\eta=0.5$.

\begin{figure}[!htbp]
\centering
\includegraphics[width=\linewidth]{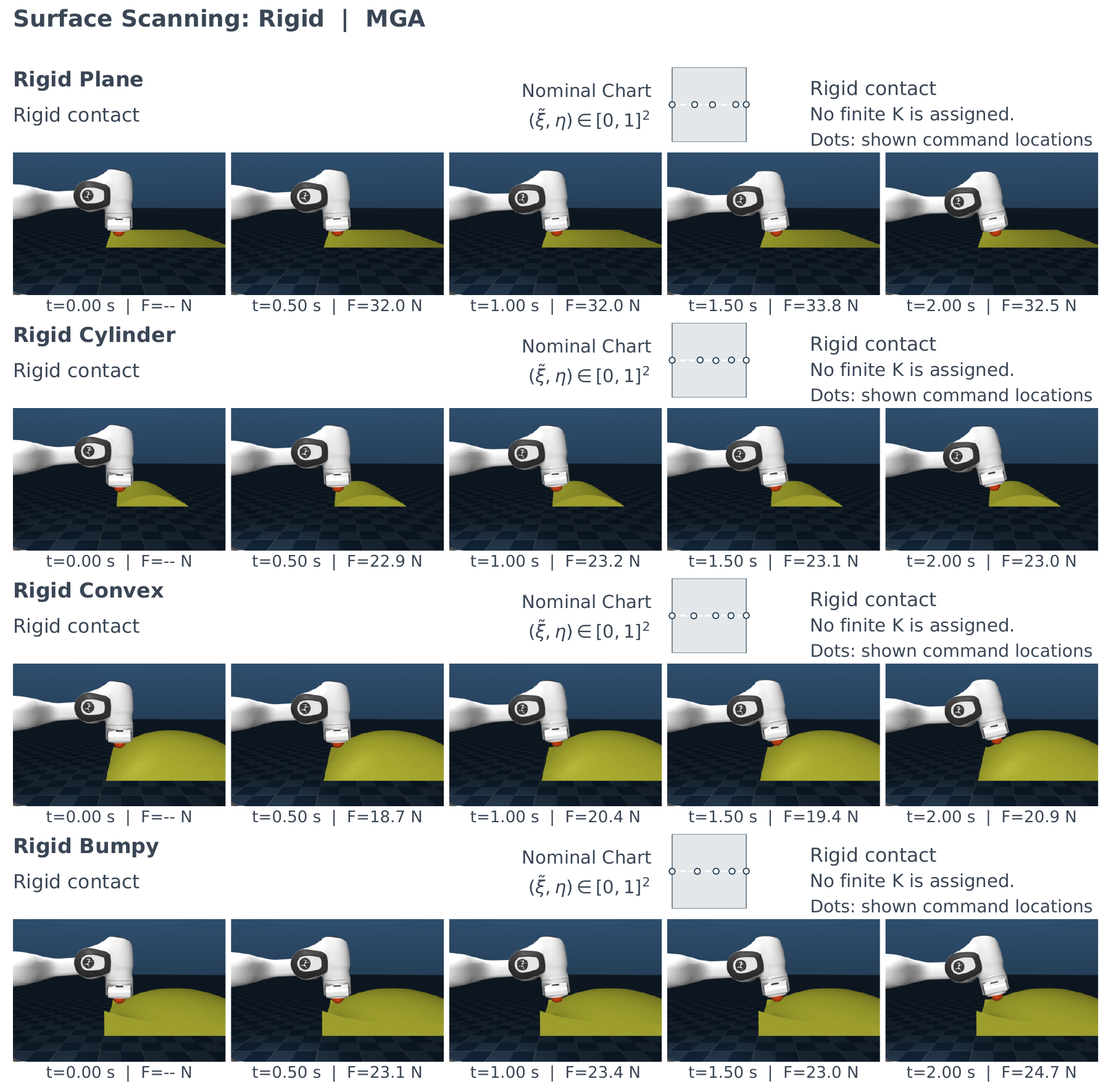}
\caption{MGA scans the four rigid surfaces: plane, cylinder, convex, and bumpy.
Five recorded states per row show the executed robot and tool
motion from the same seed, with a fixed camera and common rendering style.
The contact inset identifies the rigid regime without a finite-stiffness
color assignment.}
\label{fig:app_vis_scan_hard}
\end{figure}

\begin{figure}[!htbp]
\centering
\includegraphics[width=\linewidth]{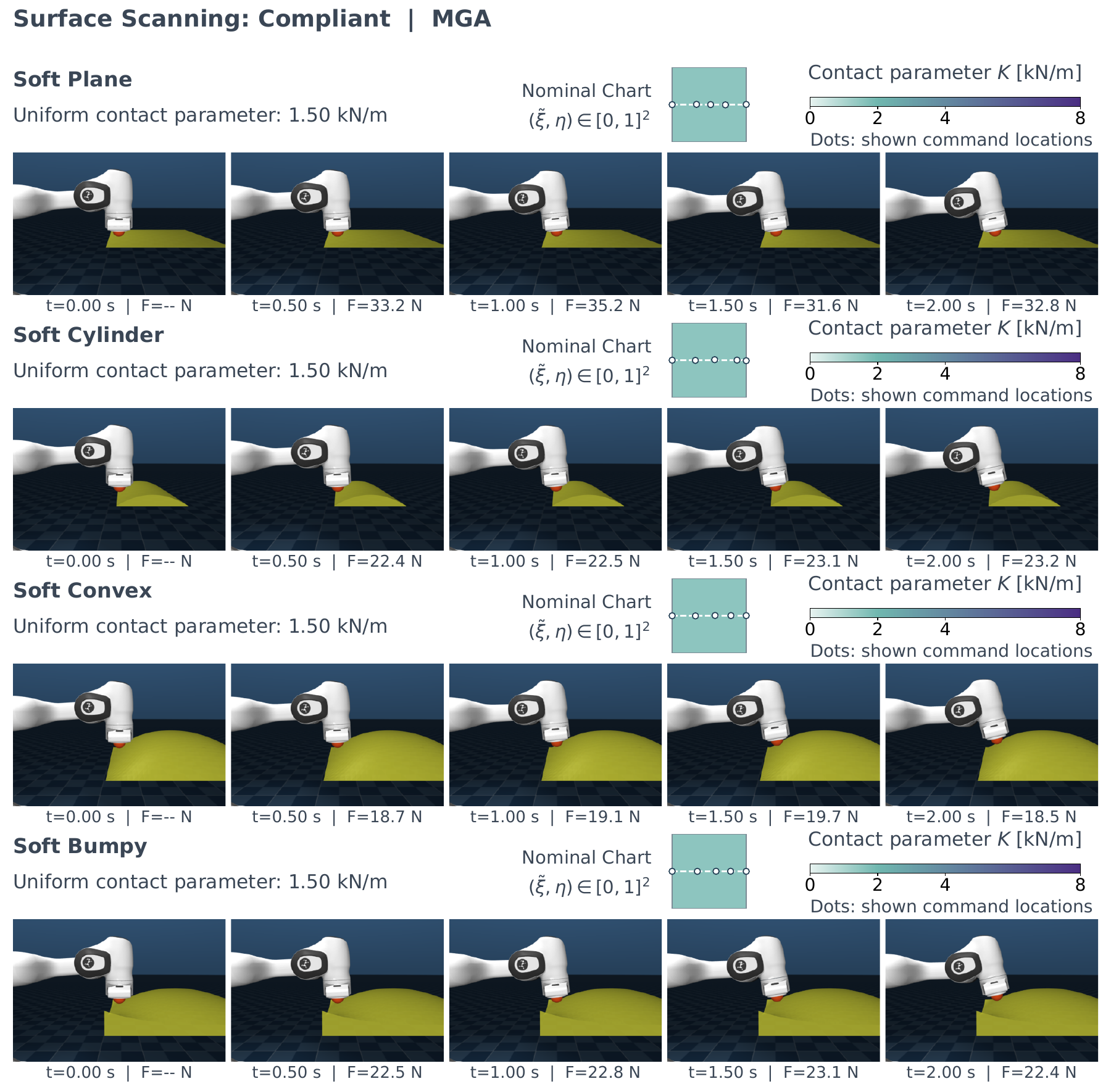}
\caption{MGA scans the four compliant surfaces. The frames show
recorded robot motion against the nominal rendered surface. Mesh appearance
alone does not measure deformation, which is quantified separately in
Fig.~\visualref{fig:app_scan_tradeoff}. Contact-parameter insets distinguish
the compliant setting from the corresponding rigid geometry.}
\label{fig:app_vis_scan_soft}
\end{figure}

\begin{figure}[!htbp]
\centering
\includegraphics[width=\linewidth]{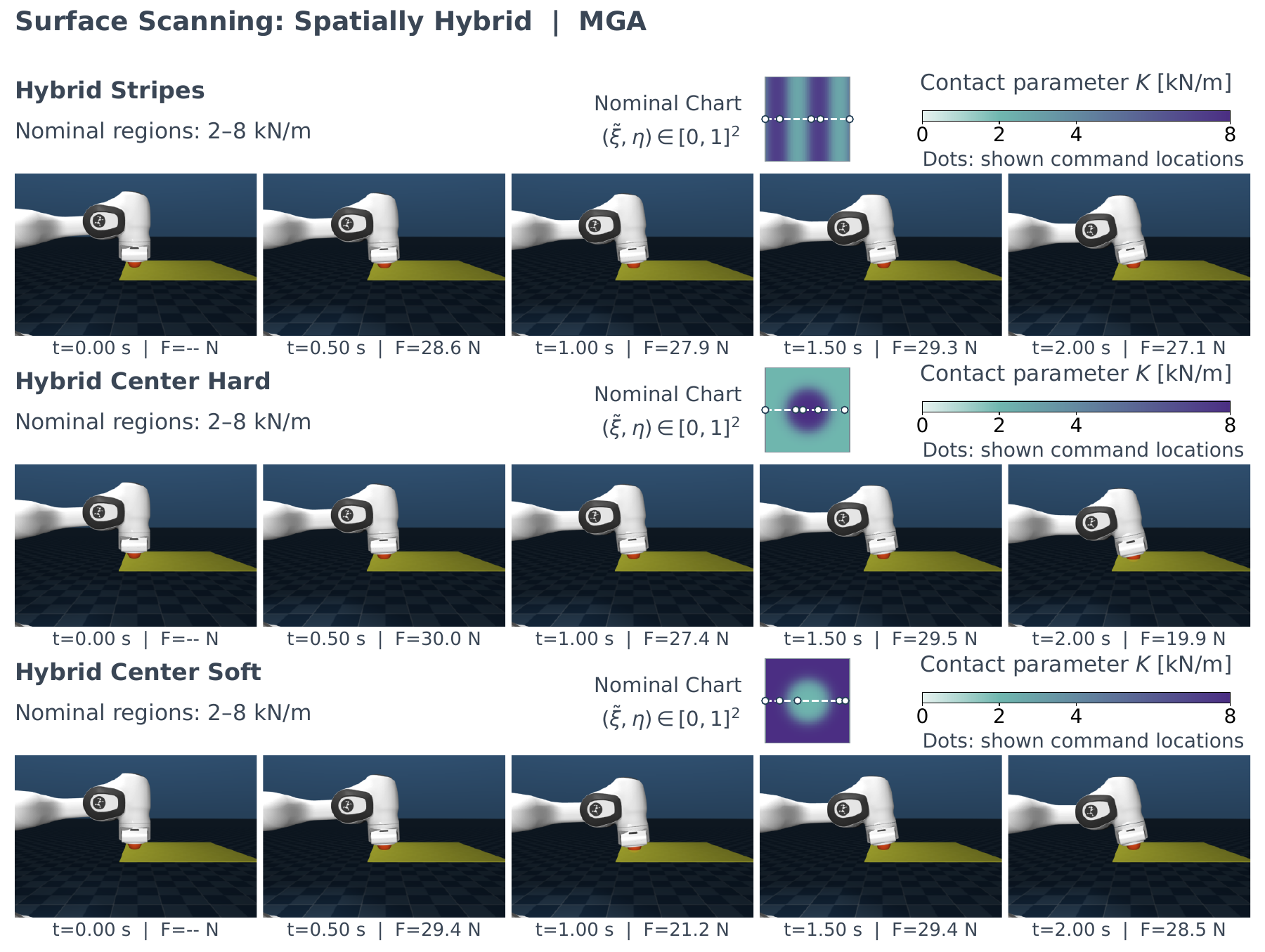}
\caption{The frames show MGA executions across three planar hybrid stiffness maps:
alternating stripes, hard center, and soft center. Material variation changes the contact
response without introducing a curved surface. Insets show the nominal
task-chart stiffness field. It is evaluated at the commanded surface
coordinate, not inferred from the displayed end-effector position.
All frames are states from the recorded executions.}
\label{fig:app_vis_scan_hybrid}
\end{figure}

\begin{figure}[!htbp]
\centering
\includegraphics[width=\linewidth]{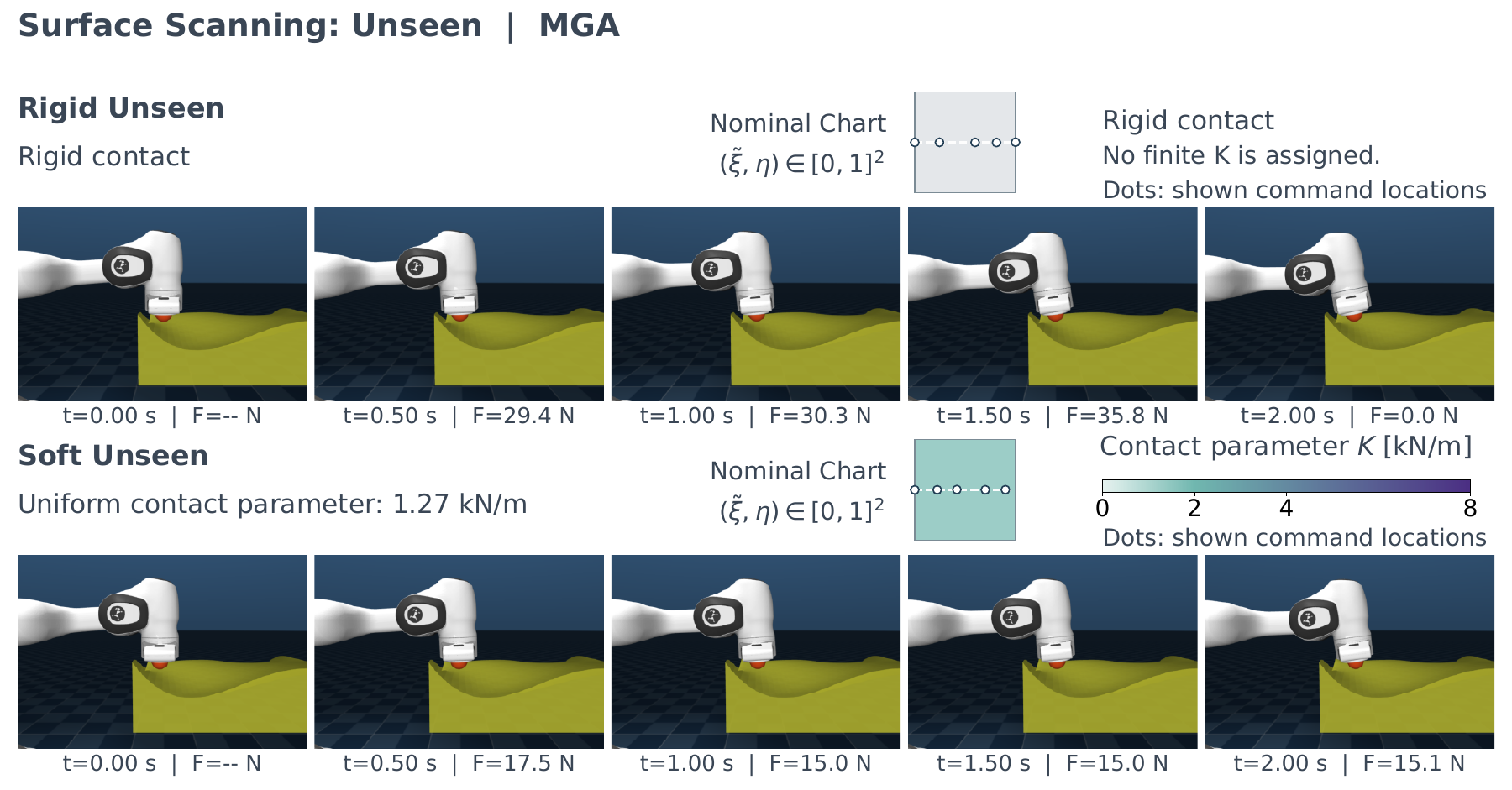}
\caption{MGA scans the held-out rigid and compliant surfaces.
The rigid example is retained despite failing the scanning success criterion,
so visual coverage of the benchmark does not select only successful cases.}
\label{fig:app_vis_scan_unseen}
\end{figure}

\clearpage
\subsection{Peg Insertion}
\label{app:vis_peg}

\begin{figure}[H]
\centering
\includegraphics[width=\linewidth]{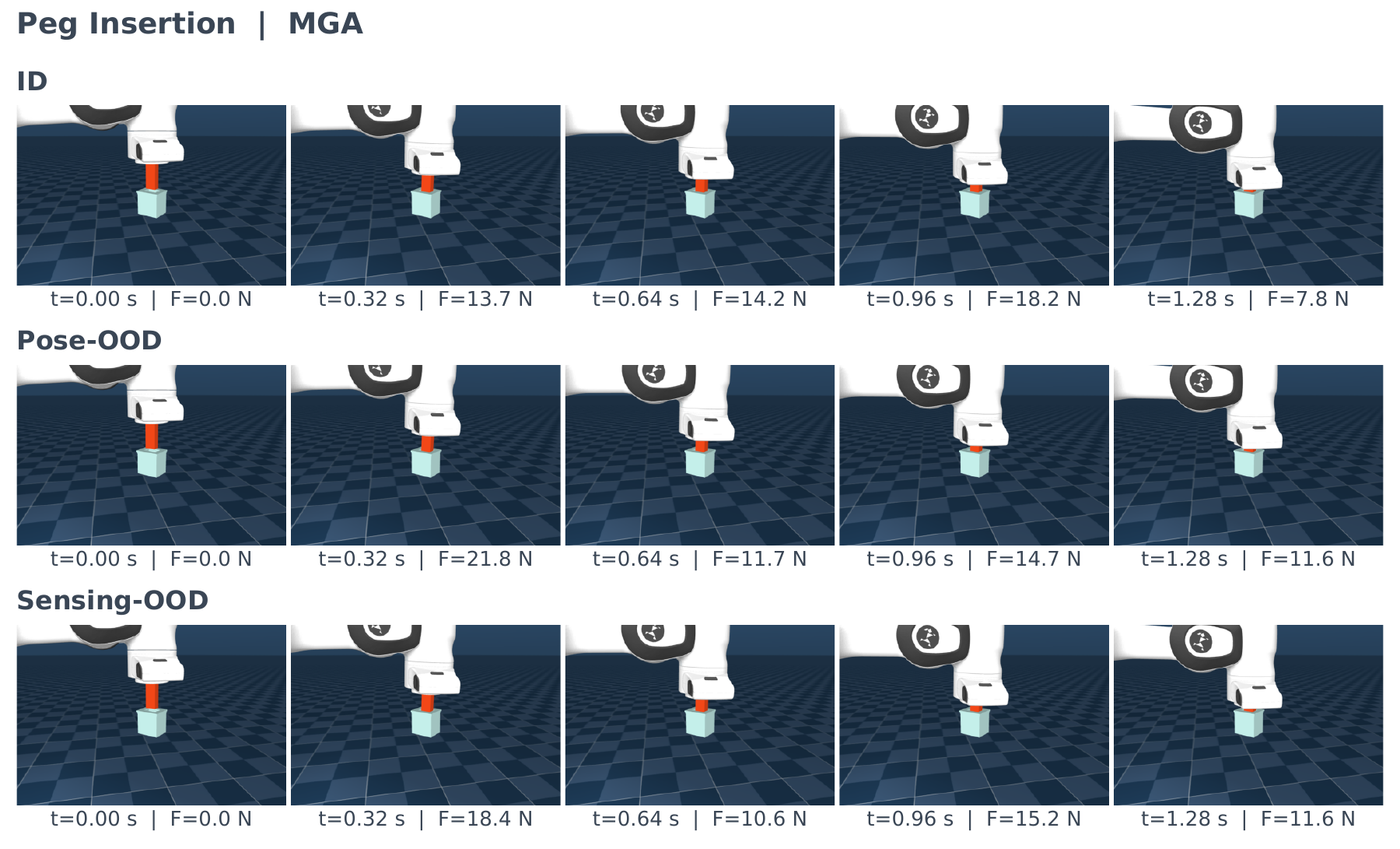}
\caption{Five chronological pose replays show \MGA{} under ID, Pose-OOD, and
Sensing-OOD. Native robot views replay recorded joint configurations without
re-simulation. Quantitative plots use the saved task-coordinate signals.
The fixed examples all complete safely. The statistical comparisons
retain every trial, including the two unsafe Pose-OOD executions.}
\label{fig:app_vis_peg}
\end{figure}

\subsection{Humanoid Push}
\label{app:vis_humanoid}

\begin{figure}[H]
\centering
\includegraphics[width=\linewidth]{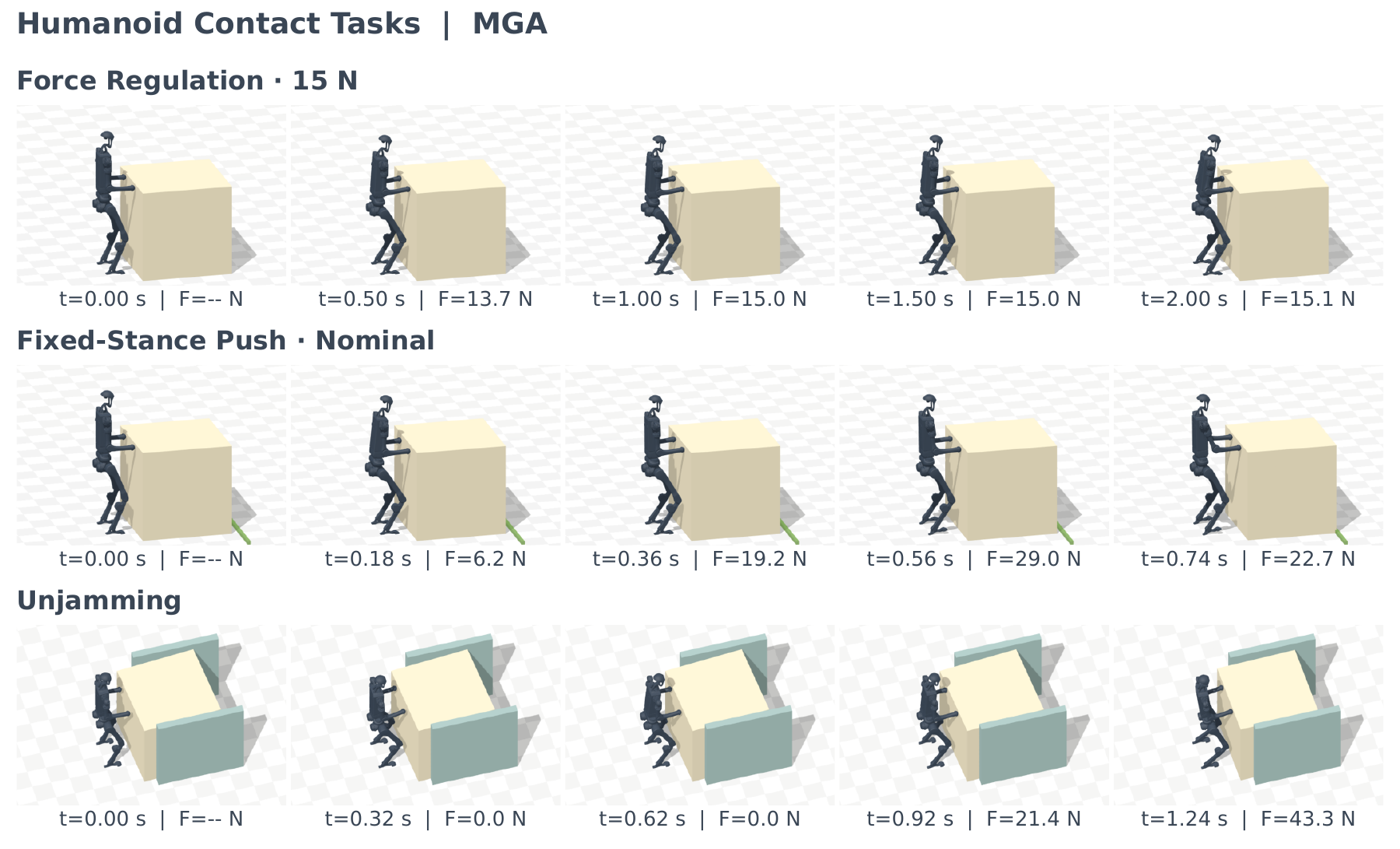}
\caption{The frames show MGA in Force Regulation, Fixed-stance Push, and Unjamming.
Five frames span each execution before terminal padding. Goal markings
appear only in the corresponding displacement tasks. These are not walking pushes.}
\label{fig:app_vis_humanoid}
\end{figure}
\endgroup

\end{document}